\documentclass[10pt,journal,compsoc]{IEEEtran}
\usepackage{amsmath,amsfonts}
\usepackage{algorithmic}
\usepackage{algorithm}
\usepackage{array}
\usepackage[caption=false,font=normalsize,labelfont=sf,textfont=sf]{subfig}
\usepackage{textcomp}
\usepackage{stfloats}
\usepackage{url}
\usepackage{verbatim}
\usepackage{graphicx}

\usepackage{amsmath}
\usepackage{booktabs} 
\usepackage{graphicx} 
\usepackage{multirow}
\usepackage{makecell} 
\usepackage{array}
\usepackage{adjustbox}
\usepackage{tabularx}
\usepackage{graphicx}
\usepackage{amssymb} 
\usepackage{pifont} 
\usepackage{adjustbox}
\usepackage[pagebackref=true,breaklinks=true,colorlinks,bookmarks=false]{hyperref}

\newcommand{\aupro}{\mbox{AU-PRO}}
\newcommand{\auprothirty}{\mbox{AU-PRO\textsubscript{0.30}}}
\newcommand{\auprofive}{\mbox{AU-PRO\textsubscript{0.05}}}
\newcommand{\mad}{\mbox{MVTec\hspace{1.5pt}AD}}
\newcommand{\madtwo}{\mbox{MVTec\hspace{1.5pt}AD\hspace{1.5pt}2}}
\newcommand{\testpublic}{\mbox{\textit{TEST\textsubscript{pub}}}}
\newcommand{\testprivate}{\mbox{\textit{TEST\textsubscript{priv}}}}
\newcommand{\testprivatemixed}{\mbox{\textit{TEST\textsubscript{priv,mix}}}}
\def\sup#1{$^{#1}$}
\newcommand{\xmark}{\ding{55}}
\newcommand{\cmark}{\ding{51}} 

\usepackage[capitalize]{cleveref}
\crefname{section}{Sec.}{Secs.}
\crefname{section}{Section}{Sections}
\crefname{table}{Table}{Tables}
\crefname{table}{Tab.}{Tabs.}

\begin{document}

\title{The MVTec AD 2 Dataset: Advanced Scenarios for Unsupervised Anomaly Detection}

\author{Lars Heckler-Kram\sup{1,2}, Jan-Hendrik Neudeck\sup{1}, Ulla Scheler\sup{1}, Rebecca König\sup{1}, Carsten Steger\sup{1} \\
  \scriptsize{\sup{1}MVTec Software GmbH, Germany\\
  \sup{2}Technical University of Munich, Germany\\}
\thanks{\copyright~2025 IEEE. Personal use of this material is permitted. Permission
from IEEE must be obtained for all other uses, in any current or future
media, including reprinting/republishing this material for advertising or
promotional purposes, creating new collective works, for resale or
redistribution to servers or lists, or reuse of any copyrighted
component of this work in other works.}}

\markboth{}%
{Heckler-Kram \MakeLowercase{\textit{et al.}}: The MVTec AD 2 Dataset: Advanced Scenarios for Unsupervised Anomaly Detection}

\maketitle

\begin{abstract}
In recent years, performance on existing anomaly detection benchmarks like \mad\ and VisA has started to saturate in terms of segmentation \aupro, with state-of-the-art models often competing in the range of less than one percentage point. This lack of discriminatory power prevents a meaningful comparison of models and thus hinders progress of the field, especially when considering the inherent stochastic nature of machine learning results. We present \madtwo, a collection of eight anomaly detection scenarios with more than 8000 high-resolution images. It comprises challenging and highly relevant industrial inspection use cases that have not been considered in previous datasets, including transparent and overlapping objects, dark-field and back light illumination, objects with high variance in the normal data, and extremely small defects. We provide comprehensive evaluations of state-of-the-art methods and show that their performance remains below 60\% average \aupro\@. Additionally, our dataset provides test scenarios with lighting condition changes to assess the robustness of methods under real-world distribution shifts. We host a publicly accessible evaluation server that holds the pixel-precise ground truth of the test set (\url{https://benchmark.mvtec.com/}). All image data is available at \url{https://www.mvtec.com/company/research/datasets/mvtec-ad-2}. 
\end{abstract}


\section{Introduction}

\begin{figure*}[t]
\centering
  \includegraphics[width=\textwidth]{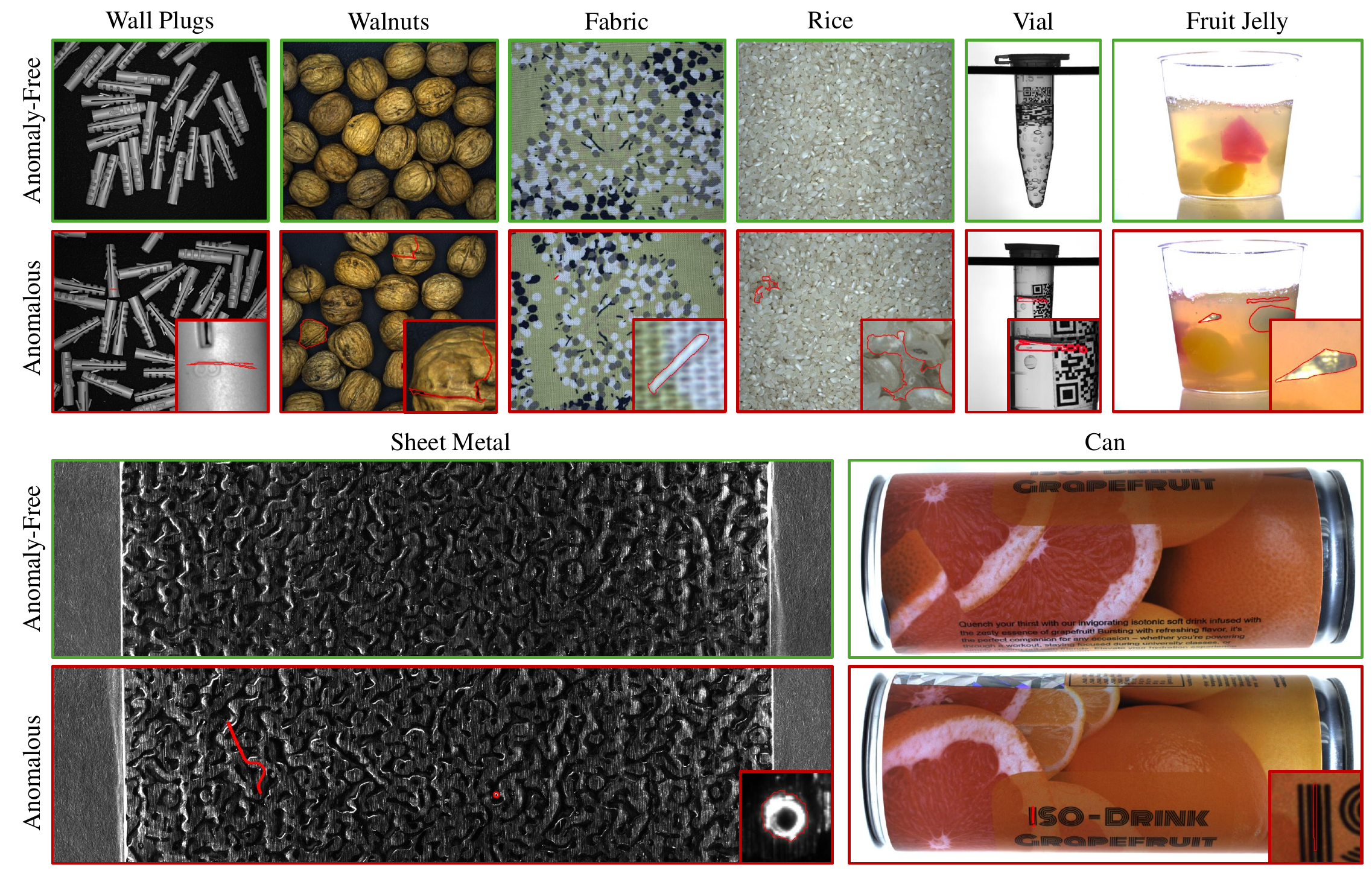}
\caption[\madtwo\ Objects Overview]{The \madtwo\ objects. For each object, one defect-free image and one image with anomalies, outlined in red, is shown. The close-up of the anomaly region shows the pixel-precise ground truth labels.}
\label{fig:mad2_overview}
\end{figure*}

\IEEEPARstart{W}{ithin} the domain of computer vision, the task of unsupervised anomaly detection and localization has gained significant attention over the past few years.
Identifying anomalous data at test time while having access only to anomaly-free data at training time is a challenge for researchers from various fields such as autonomous driving \cite{blum2019_fishyscapes_dataset,hendrycks2019_benchmark_anomaly_segmentation_autodriving}, healthcare \cite{seebock2019_uncertainty_for_ad,menze2015_brats_dataset}, video surveillance \cite{nazare2018_pretrained_cnns_for_ad,li2013_ucsd_video_ad_dataset}, and visual inspection \cite{bergmann2019_mvtec_ad_cvpr,Zou_2022_VisA}.
Especially in industrial settings, visual anomaly detection serves as an important tool for quality assurance.
Here, anomaly detection systems not only are expected to classify images of manufactured goods as \textit{reject} or \textit{accept} but also to provide pixel-precise anomaly maps that correspond to the anomalous regions.
The reliable localization of defects ensures the interpretability and trustworthiness of the system, enables an early detection of systematic malfunctions within the production line, and enables an informed decision about the severity of the defect during post-processing.

The progress of the field of industrial anomaly detection is mainly driven by the availability of suitable datasets.
Early datasets such as \mad\ \cite{bergmann2019_mvtec_ad_cvpr,bergmann2021_mvtec_ad_ijcv} or VisA \cite{Zou_2022_VisA} focused their efforts on providing a wide range of industrial anomaly detection scenarios, defining the task, and setting standards in evaluation metrics and benchmark protocols.
Later, anomaly detection datasets with new types of anomalies such as logical constraints \cite{bergmann2021_mvtec_loco_ijcv} or additional modalities such as 3D information \cite{bergmann2022_mvtec_3d_ad, bonfiglioli_eyecandies_2022} were introduced.
Although the full spectrum of industrial visual inspection settings, e.g., different types of illumination, is not yet covered by existing datasets, from a practical point of view, this collaborative effort of the research community has enabled the development of a large number of methods that can now be deployed in real-world production systems.
Simultaneously, established benchmarks, e.g., \mad\ or VisA, begin to saturate. 
Often, the performance of state-of-the-art models differs by less than one percentage point.
Because of the stochastic nature of deep learning models and their dependence on well-chosen hyperparameters, this makes it challenging to distinguish truly innovative approaches from incremental improvements.

Moreover, the lack of consistent evaluation settings hinders a meaningful comparison of models.
Although metrics such as segmentation \aupro\ \cite{bergmann2021_mvtec_ad_ijcv} are well-defined, occasionally slightly adapted or approximated versions are used.
What is worse, the common practice of tuning hyperparameters based on test data imply the inappropriate use of the test set since this contradicts the foundational definition of unsupervised anomaly detection, i.e., not knowing which defects to expect.

\begin{table}[ht]
\centering
\caption{Anomaly segmentation \auprothirty\ (in \%) and mean performance of state-of-the-art methods on established anomaly detection datasets and our \madtwo\ dataset. While segmentation performance of state-of-the-art methods saturates on established datasets, \madtwo\ introduces new challenges and opportunities for genuine methodological improvements.}
\label{tab:performance_on_other_ad_datasets}
\begin{adjustbox}{max width = .98\columnwidth}
\begin{tabular}{cccc} 
\toprule
\multicolumn{1}{l}{Dataset} & \multicolumn{1}{l}{\mad\ \cite{bergmann2019_mvtec_ad_cvpr}} & \multicolumn{1}{l}{VisA \cite{Zou_2022_VisA}} & \multicolumn{1}{l}{\madtwo (ours)}  \\
\midrule
PatchCore                   & 92.7                                                       & 79.7                                          & 53.8                     \\
RD                          & 93.9                                                       & 89.0                                          & 53.0                      \\
RD++                        & 95.0                                                       & 89.3                                          & 54.3                      \\
EfficientAD                 & 93.5                                                       & 94.0                                          & 58.7                      \\
MSFlow                      & 91.2                                                       & 88.4                                          & 52.7                      \\
SimpleNet                   & 89.6                                                       & 68.9                                          & 46.4                      \\
DSR                         & 90.8                                                       & 68.1                                          & 49.0                      \\
\midrule\midrule
Mean                        & 92.4                                                       & 82.4                                          & 52.6                      \\
\bottomrule
\end{tabular}
\end{adjustbox}
\end{table}

\begin{table*}[ht]
\centering
\caption{Qualitative comparison of datasets based on various criteria. \madtwo\ provides real-world images with high variability in their anomaly-free state as well as different lighting conditions to investigate model behavior under distribution shifts.  Another main contribution, omitted here, is its considerable difficulty (\Cref{tab:performance_on_other_ad_datasets}). \madtwo\ does not focus on logical anomalies or 3D data.}\label{tab:dataset_features}
\begin{adjustbox}{max width = .98\linewidth}
\begin{tabular}{@{}lcccccc|c@{}}
\toprule 
 &           \thead{MVTec\\AD}  & \thead{MVTec\\LOCO AD}& VisA         & \thead{MVTec\\3D AD}  & \thead{Eye-\\candies}   & \thead{Real-\\IAD}     & \thead{MVTec AD 2\\(ours)} \\ \midrule
Real images                                 & \cmark & \cmark & \cmark & \cmark & \xmark     & \cmark & \cmark \\
High Variability*                 & \xmark     & \cmark & \xmark     & \cmark & \cmark & \xmark     & \cmark \\
Defects at Image Borders                    & \xmark     & \xmark     & \xmark     & \xmark     & \xmark     & \xmark     & \cmark \\
Variation in \#Objects/image*               & \xmark     & \xmark     & \xmark     & \xmark     & \xmark     & \xmark     & \cmark \\
Structural Defects                          & \cmark & \cmark & \cmark & \cmark & \cmark & \cmark & \cmark  \\
Logical Defects                             & \xmark     & \cmark & \xmark     & \xmark     & \xmark     & \xmark     & \xmark\\
2D Data                                     & \cmark & \cmark & \cmark & \cmark & \cmark & \cmark & \cmark  \\
3D Data                                     & \xmark     & \xmark     & \xmark     & \cmark & \cmark & \xmark     & \xmark \\
Different Lighting Conditions               & \xmark     & \xmark     & \xmark     & \xmark     & \cmark & \xmark     & \cmark \\
Transparent Objects                         & \xmark     & \xmark     & \xmark     & \xmark     & \cmark & \xmark     & \cmark \\ \bottomrule
\multicolumn{8}{l}{\footnotesize{*Refers to the allowed variation in the normal data used during training and validation}}\\
\end{tabular}
\end{adjustbox}
\end{table*}

To contribute to the advancement of anomaly detection research, we present \madtwo, an anomaly detection dataset and benchmark containing eight advanced anomaly detection scenarios (\Cref{fig:mad2_overview}).
\madtwo{} provides a significant challenge for current state-of-the-art methods compared to existing datasets (\Cref{tab:performance_on_other_ad_datasets}).
Methods that achieve over 90\% \auprothirty\ on established datasets like \mad\ score a maximum of 58.7\% on \madtwo. 
Our new dataset focuses on settings that are highly relevant in industrial inspection and have been neglected by existing datasets (\Cref{tab:dataset_features}).
It includes small defects in large images, defects at the image border, object categories with high variability in their anomaly-free state, transparent and reflective objects, and bulk goods where object instances might overlap and vary randomly in position and quantity.
In addition, \madtwo\ expands current anomaly detection benchmarks in two important ways: 
First, every single scene was captured under at least four different lighting conditions.
Consequently, the developed dataset design offers new possibilities to evaluate the robustness of anomaly detection methods against real-world distribution shifts due to illumination changes.
These naturally occur in industrial applications where one model is deployed on different machines, as well as due to aging of the illumination.

Second, since ground truth evaluation is exclusively possible on our publicly available benchmark server, the optimization on the test set is prevented.
In this manner, \madtwo\ paves the way for a standardized and, therefore, fairer comparison between different approaches.
Overall, the contribution of our work is threefold:
\begin{itemize}
    \item We present \madtwo, a novel and highly challenging dataset, on which the current best model achieves only 58.7\% \auprothirty. This opens up opportunities for significant improvements within the field of visual anomaly detection.
    \item For the first time, we enable assessing the robustness of anomaly detection methods under real-world distribution shifts induced by lighting condition changes tailored to the respective object scenarios.
    \item We provide a public evaluation server for the research community to measure the progress in the field of anomaly detection more reliably. In addition, we evaluate seven state-of-the-art methods on \madtwo\ and discuss their real-world applicability.
\end{itemize}

\section{Related work}
Industrial visual quality control relies on anomaly detection systems to reliably localize, i.e., segment, anomalous regions.
To evaluate these methods quantitatively, datasets with pixel-precise segmentation ground truth are essential. 
This section provides a brief review of the most commonly used anomaly detection datasets that feature pixel-precise anomaly mask labels.
Additionally, an overview of existing anomaly detection methods is presented.

\subsection{Anomaly Detection Datasets}
\mad\ \cite{bergmann2019_mvtec_ad_cvpr} consists of 5,354 high-resolution color images covering 15 distinct object categories.
While this dataset has accelerated research for industrial anomaly detection, today performance on it nearly saturates with the best model achieving 97.8\% mean \aupro\ \cite{cpr}.
Similarly, VisA \cite{Zou_2022_VisA} contains 10,821 images from 12 object categories, including four multi-object settings with a fixed number of samples per image. It expands the number of standard anomaly detection scenarios available for scientific research. Similar to \mad, VisA is close to being solved \cite{batzner2023_efficientad, zhang2023_diffusionad}.
MVTec LOCO AD \cite{bergmann2021_mvtec_loco_ijcv} is a dataset that focuses on logical anomalies, where elements in a defective image maintain the same general appearance but violate logical constraints. For example, this might include an incorrect number of elements in a box.
Here, the currently best model reaches 79.8\% in localization performance \cite{batzner2023_efficientad}.
Real-IAD \cite{wang_real-iad_2024} is a more recent dataset that focuses on multiple viewing angles and a larger set of non-deformable objects captured as single samples, including object types that were seen in previous datasets, such as \mad\ and VisA.
The dataset is notable for its extensive size, featuring over 150,000 high-resolution images of 30 different objects. However, the anomaly-free conditions of the objects show little variation. Besides, in the unsupervised setting, existing models already reach a pixel \auprothirty\ of 94.4\%, as outlined in the original work \cite{wang_real-iad_2024,deng2022_rd}, even at a resolution of 1/20 of the original image size. This suggests that defects are well distinguishable from the distribution of normal data.
Other datasets opened up new domains in anomaly detection, such as the extension to 3D data \cite{bergmann2022_mvtec_3d_ad} or multi-pose \cite{zhou2023_pad}. 
The Eyecandies dataset \cite{bonfiglioli_eyecandies_2022} presents synthetically generated data of ten different kinds of sweets, including transparent and asymmetric objects.
In addition to providing perfect ground truth, synthetic rendering offers the advantage of simulating different lighting conditions. 

We conclude that there are two main reasons current datasets are not conducive to new research: First, existing datasets are too saturated to lead to much differentiation between models.
Recent new datasets, although they cover new problem settings, essentially replicate existing difficulty levels in terms of the objects they use and thus were already more than 90\% solved at the time of their publication.
Second, although the aforementioned datasets cover a wide range of applications for unsupervised visual anomaly detection, we observed that essential real-world use cases are missing within this collection.
To name just one example, production rates of modern machines might necessitate the visual inspection of samples in bulk, where the sample positions and quantity vary greatly.
Moreover, in practice, an anomaly detection system might need to be deployed on several machines with slightly different environmental conditions, e.g., lighting, without retraining.
With \madtwo, we present a collection of eight 2D anomaly detection scenarios that is sufficiently difficult and versatile to develop methods suitable for solving these challenges based on real-world data.

\subsection{Unsupervised Anomaly Detection Methods}
Owing to the rigorous work of researchers in the past few years, various approaches for tackling the task of unsupervised anomaly detection exist.
Reconstruction-based methods \cite{bergmann2018_ssim_ae,Zavrtanik_DRAEM,zhang2023_diffad} recover the anomaly-free state of an image and compute the differences between input and reconstruction.
Other approaches memorize the distribution of normal data either explicitly inside memory banks \cite{patchcore} or implicitly, for example, by storing statistical properties of the embeddings of anomaly-free data \cite{cohen2020_spade,padim} or mapping them to a predefined distribution using normalizing flows \cite{Gudovskiy_2022_CFlow,csflow,fastflow,zhou2024_msflow}.
Student--teacher methods \cite{bergmann2020_uninformed_cvpr,asymST,batzner2023_efficientad,deng2022_rd,tien2023_rd_revisited} combine both principles to a certain extent.
Here, a student network learns to imitate a pretrained teacher network for the defect-free training data.
At test time, the presence of anomalies results in deviations between the two model outputs.
Leveraging the power of large language models, embeddings of text describing the normal state and the test image can also be compared to detect anomalies \cite{jeong2023_winclip,gu2024_anomalygpt}.
Besides, some frameworks apply synthetic anomalies to discriminate the distributions of anomaly-free and anomalous data \cite{Schlueter_2022_nsa,Zavrtanik_2022_DSR,Liu_2023_simplenet}.

Since current anomaly detection benchmarks such as \mad\ \cite{bergmann2019_mvtec_ad_cvpr} or VisA \cite{Zou_2022_VisA} approach saturation, the development of new methods for visual anomaly detection is hindered. Additionally, genuine progress is difficult to discern from incremental improvements when the performance differences of state-of-the-art models fall within the range of stochastic deviations.
This calls for new challenges, especially in the standard setting of 2D visual inspection.
Our proposed dataset \madtwo\ provides advanced anomaly detection scenarios to stimulate the design of new approaches and to enable a fairer and more meaningful comparison of methods.

\section{The MVTec AD 2 dataset}
\label{subsec:dataset_description}
\subsection{Advanced Anomaly Detection Scenarios}
\madtwo\ comprises eight challenging scenarios for unsupervised anomaly detection in industrial applications, featuring a total of 8,004 high-resolution images.
\Cref{fig:mad2_overview} presents sample images depicting both defect-free and anomalous object instances.
Each scenario was selected to highlight specific difficulties. 
The bulk goods scenarios \textit{(Wall Plugs, Walnuts, Rice)} involve overlapping and occluded objects arranged in uncontrolled spatial patterns. 
Textured objects \textit{(Fabric, Sheet Metal)} were chosen for the high variability in their normal data. Reflective metal objects \textit{(Sheet Metal, Can)} and transparent objects \textit{(Vial, Fruit Jelly)} introduce additional challenges due to reflections and distortions, respectively.
Defects vary by type and cover surface imperfections like scratches and holes, contamination by foreign bodies, and missing parts. \Cref{tab:scenario_overview} summarizes the main challenges of each scenario that were derived from a variety of real-world applications.

\begin{figure}[ht]
\centering
\begin{adjustbox}{max width = \columnwidth}
  \includegraphics[]{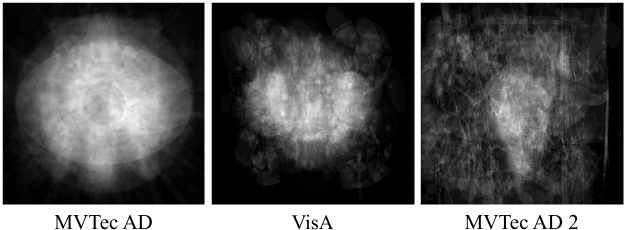}
\end{adjustbox}
\caption{Defect distribution across the \mad, VisA, and \madtwo\ datasets. The plot shows the normalized ground truth label distribution for each dataset. In \mad\ and VisA, defect labels are predominantly concentrated at the center of the images, whereas \madtwo\ exhibits a significant number of defects at the image borders. This enables to test methods for robustness against boundary artifacts.}
\label{fig:defect_distribution}
\end{figure}

Furthermore, we designed the anomalies such that they are distributed across the entire height and width of the images and also appear directly at the image borders (\Cref{fig:defect_distribution}). This poses new challenges for models trained with center cropping and allows for an evaluation of frequently occurring padding artifacts.

To capture the diverse spectrum of industrial visual inspection, in addition to RGB images, the dataset includes single-channel gray-value images (\textit{Sheet Metal}, \textit{Vial}, \textit{Wall Plugs}).
Furthermore, the dataset features illuminations during object acquisition that are commonly used in industrial practice but neglected in existing datasets, e.g., back-light illumination \textit{(Vial, Fruit Jelly)} and dark-field illumination \textit{(Sheet Metal)} \cite[Chapter~2.1.5]{steger:18}.
Image resolutions range from 2.6 to 5 megapixels with varying aspect ratios, as detailed in \Cref{tab:statistical_overview}.

\begin{table*}[ht]
  \centering
  \caption{Acquisition setting, description of occurring defects, and main challenges associated with each \madtwo\ object.}
  \label{tab:scenario_overview}
  \renewcommand{\arraystretch}{1.5}
\begin{adjustbox}{max width = .98\linewidth}
\begin{tabularx}{\textwidth}{>{\setlength\hsize{0.11\hsize}\setlength\linewidth{\hsize}}X>{\setlength\hsize{.17\hsize}\setlength\linewidth{\hsize}\raggedright}X>{\setlength\hsize{.32\hsize}\setlength\linewidth{\hsize}\raggedright\arraybackslash}X>{\setlength\hsize{0.4\hsize}\setlength\linewidth{\hsize}\raggedright\arraybackslash}X}
    \toprule
    Object & Acquisition Setting & \multicolumn{1}{c}{Description of Occurring Defects} & \multicolumn{1}{c}{Main Challenges} \\ 
    \midrule
    \midrule
Can & diffuse bright-field front light illumination of soda can &
\begin{minipage}[t]{\linewidth}
\begin{itemize}
\item print defects
\item scratches
\end{itemize}
\end{minipage} &
\begin{minipage}[t]{\linewidth}
\begin{itemize}
\item tiny defects on a textured surface
\item light reflections
\item high variability in the normal data due to rotation of the cans
\item images with large aspect ratio
\end{itemize}
\end{minipage}\\  
\midrule
Fabric & diffuse bright-field front light illumination for fabric inspection &
\begin{minipage}[t]{\linewidth}
\begin{itemize}
\item cuts and holes
\item color inconsistencies
\item loose threads and extra fabric pieces
\end{itemize}
\end{minipage} &
\begin{minipage}[t]{\linewidth}
\begin{itemize}
\item tiny defects with low contrast
\item high variability of the normal data due to varying pattern of the print
\end{itemize} 
\end{minipage}\\
\midrule
Fruit Jelly & diffuse bright-field back and front light illumination of jelly with pieces of fruit &
\begin{minipage}[t]{\linewidth}
\begin{itemize}
\item foreign object contamination of different texture and size
\item surface scratches
\end{itemize}
\end{minipage} &
\begin{minipage}[t]{\linewidth}
\begin{itemize}
\item high variability of the normal data due to varying appearance and amount of ingredients
\item semi-transparency
\item low-contrast defects
\end{itemize} 
\end{minipage}\\
\midrule
Rice & diffuse bright-field front light illumination of rice grains in bulk &
\begin{minipage}[t]{\linewidth}
\begin{itemize}
\item contamination by (semi-)trans-parent plastic
\item contamination by foreign objects
\end{itemize}
\end{minipage} &
\begin{minipage}[t]{\linewidth}
\begin{itemize}
\item very low contrast defects
\item transparent contaminants
\end{itemize} 
\end{minipage}\\
\midrule
Sheet Metal & directed dark-field front light illumination of pieces of sheet metal &
\begin{minipage}[t]{\linewidth}
\begin{itemize}
\item scratches, cuts, and holes
\item foreign objects
\end{itemize}
\end{minipage} &
\begin{minipage}[t]{\linewidth}
\begin{itemize}
\item random surface reflections
\item images with large aspect ratio
\item multiple defects of varying size per image
\end{itemize} 
\end{minipage}\\
\midrule
Vial & diffuse bright-field back light illumination of a single vial with clear sparkling liquid &
\begin{minipage}[t]{\linewidth}
\begin{itemize}
\item contamination by foreign objects
\item damaged or missing QR codes
\item fill level deviations
\item open or missing lids
\end{itemize}
\end{minipage} &
\begin{minipage}[t]{\linewidth}
\begin{itemize}
\item transparency of vials and defects
\item small defects with low contrast
\item high variability due to rotation of vials, air bubbles, and QR codes
\end{itemize} 
\end{minipage}\\
\midrule
Wall Plugs & diffuse bright-field front light illumination of multiple wall plugs &
\begin{minipage}[t]{\linewidth}
\begin{itemize}
\item scratches, cuts, and missing parts
\item contamination by foreign objects or broken pieces
\item incorrect size of wall plugs
\end{itemize}
\end{minipage} &
\begin{minipage}[t]{\linewidth}
\begin{itemize}
\item bulk items: touching, overlapping, and partially cut-off by the image frame
\item varying quantity and placement of plugs
\end{itemize} 
\end{minipage}\\
\midrule 
Walnuts & diffuse bright-field front light illumination of multiple walnuts &
\begin{minipage}[t]{\linewidth}
\begin{itemize}
\item cracks and holes
\item contamination by foreign objects or incomplete walnuts
\end{itemize}
\end{minipage} &
\begin{minipage}[t]{\linewidth}
\begin{itemize}
\item bulk items: touching, overlapping, and partially cut-off by the image frame
\item high variability in size, shape, and structure of individual walnuts
\item varying quantity and placement of walnuts
\end{itemize} 
\end{minipage} \\
\bottomrule
\end{tabularx}
\end{adjustbox}
\end{table*}

\begin{table*}[ht]
  \centering
  \caption{Statistical overview of the \madtwo\ dataset. For each scenario, the number of training, validation, and test images as well as the image size is shown.}
  \label{tab:statistical_overview}
\begin{adjustbox}{max width = .98\linewidth}
\renewcommand{\arraystretch}{1.2}
\begin{tabular}{lccccccc}
    \toprule
    \multirow{2}{*}{Object} & \multirow{2}{*}{\thead{\# Train \\ images}} & \multirow{2}{*}{\thead{\# Val\\ images}} & \multicolumn{4}{c}{\thead{\# Test images \textit{(without / with anomalies)}}} &  \multirow{2}{*}{\thead{image size\\ (W$\times$H)}} \\ 
    \cline{4-7}
                    &     &     & \testpublic           & \testprivate              & \testprivatemixed       & total                     &           \\ \midrule
    Can             & 412 & 46  & 162 \textit{(72/90)}  & 321 \textit{(145/176)}    & 321 \textit{(145/176)}    & 804 \textit{(362/442)}    & 2232$\times$1024 \\
    Fabric          & 387 & 43  & 156 \textit{(66/90)}  & 314 \textit{(133/181)}    & 314 \textit{(133/181)}    & 784 \textit{(332/452)}    & 2448$\times$2048 \\ 
    Fruit Jelly     & 263 & 37  & 80  \textit{(20/60)}  & 255 \textit{(71/184)}     & 255 \textit{(71/184)}     & 590 \textit{(162/428)}    & 2100$\times$1520 \\ 
    Rice            & 313 & 35  & 132 \textit{(42/90)}  & 277 \textit{(96/181)}     & 277 \textit{(96/181)}     & 686 \textit{(234/452)}    & 2448$\times$2048 \\ 
    Sheet Metal     & 137 & 19  & 114 \textit{(24/90)}  & 142 \textit{(36/106)}     & 142 \textit{(36/106)}     & 398 \textit{(96/302)}     & 4224$\times$1056 \\ 
    Vial            & 291 & 41  & 140 \textit{(35/105)} & 276 \textit{(78/198)}     & 276 \textit{(78/198)}     & 692 \textit{(191/501)}    & 1400$\times$1900 \\ 
    Wall Plugs      & 293 & 33  & 150 \textit{(60/90)}  & 232 \textit{(96/136)}     & 232 \textit{(96/136)}     & 614 \textit{(252/362)}    & 2448$\times$2048 \\ 
    Walnuts         & 432 & 48  & 150 \textit{(60/90)}  & 228 \textit{(93/135)}     & 228 \textit{(93/135)}     & 606 \textit{(246/360)}    & 2448$\times$2048 \\   
    \bottomrule
    
    \multicolumn{8}{l}{\footnotesize{\testpublic: images and corresponding ground truth available for all lighting conditions.}} \\
    \multicolumn{8}{l}{\footnotesize{\testprivate: images available, ground truth only on evaluation server, same lighting as in the train set.}} \\
    \multicolumn{8}{l}{\footnotesize{\testprivatemixed: images available, ground truth only on evaluation server, seen and unseen lighting conditions.}} \\
    \end{tabular}

\end{adjustbox}
\end{table*}

\subsection{Dataset Design According to the Unsupervised Setting}
\label{subsec:dataset_design}
\madtwo\ comprises a training, validation, and test set.
In line with the unsupervised setting, the training and validation sets contain only non-anomalous data.
The training data consists of scenes acquired with regular lighting only since illumination changes in practice will occur mainly during test time.
The test data is divided into three parts: a publicly available example test set (\testpublic) and two private test sets (\testprivate\ and \testprivatemixed). 
The public test set (\testpublic) includes a small number of normal and anomalous images with their corresponding segmentation ground truth for all lighting conditions, facilitating local testing and an initial performance estimation.
The significantly larger private test sets only have image data publicly available, while the ground truth is private and evaluations are only possible on our benchmark server (\Cref{sec:benchmark_server}). 
\testprivate\ includes images with the same lighting conditions as the training set. 
In contrast, \testprivatemixed\ depicts the same scenes as \testprivate\ but includes both seen and unseen lighting conditions, randomly selected for each test image, encompassing both normal and anomalous data. 
This setup enables a direct comparison of model behavior under real-world distribution shifts.
Data is available for download under \url{https://www.mvtec.com/company/research/datasets/mvtec-ad-2} and released under a \mbox{CC BY-NC-SA 4.0 license}.

\subsection{Distribution Shift Through Unseen Lighting}
\label{subsec:distribution_shift}
\begin{figure*}[ht]
  \centering
  \includegraphics[width=.9\textwidth]{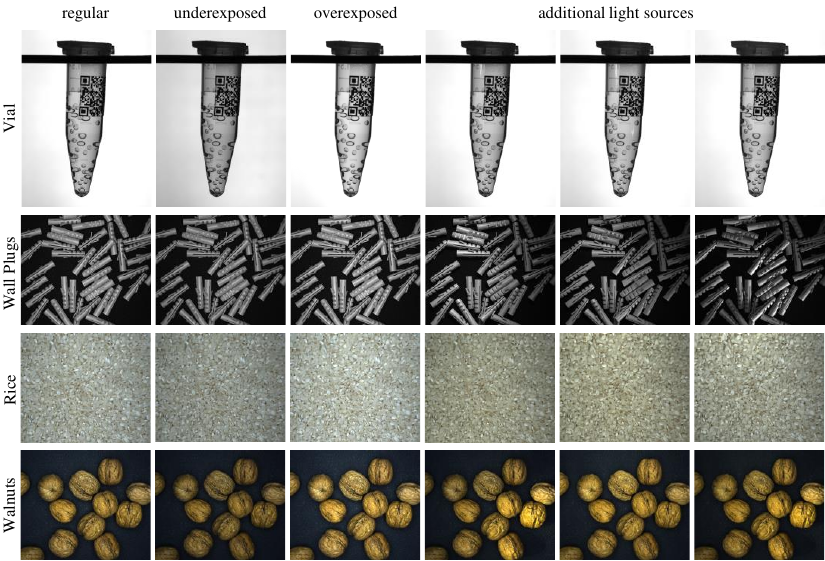}
  \caption{Lighting condition changes contained in \madtwo\ for several example objects. In addition to the regular exposure, each scene was captured under minor over- and underexposure. Moreover, additional light sources evoke object-specific variations in appearance such as reflections (\textit{Vial}), uneven illumination (\textit{Wall Plugs}), or slight changes in color temperature (\textit{Rice, Walnuts}).}
  \label{fig:lighting_conditions}
\end{figure*}

For every object, we provide images that are brighter or darker than the images seen during training by adjusting the exposure time accordingly. Further lighting shifts in the test images are tailored to the objects present in each scenario. 
For example, in the case of dark-field illumination, we used two lighting setups with different illumination angles.
For objects with reflective surfaces, additional LED spot lights created reflections.
Likewise, for bulk goods, additional spot lights led to an uneven light distribution within the scene.
For all object categories, at least four different lighting conditions were used.
\Cref{fig:lighting_conditions} shows examples of the resulting scenes.
When varying the lighting condition, all other parameters such as the position of the object or the camera parameters (apart from the exposure time) were kept constant.

\subsection{High-Quality Pixel-Precise Ground Truth} To ensure high-quality segmentation ground truth and to capture all anomalies, coarse ground truth annotations were initially created on-site during the acquisition of the images. 
These coarse labels were then refined by human annotators according to object-specific labeling guidelines. 
The refined labels underwent both manual and automated checks to minimize incomplete or spurious label regions.
We verified the pixel-precise ground truth for global consistency across different scenes and developed software tools to identify potentially mislabeled tiny regions or holes. 
Additionally, if the appearance of a defect significantly changed between lighting conditions, the annotations were adjusted accordingly. 
Finally, the anomaly-free data was manually filtered to ensure that no defects were present, thus strictly adhering to the requirements of unsupervised anomaly detection for training and validation.

\subsection{Image Acquisition Settings}
\label{subsec:acquisition_and_lighting_per_object}
In the following, the image acquisition settings from \Cref{tab:scenario_overview} are described in detail for each object.
For inducing lighting condition changes in the test data, i.e., \testpublic\ and \testprivatemixed\ (\Cref{subsec:dataset_design}), we varied the exposure time and installed additional light sources that are used in combination with the regular lighting if not explicitly stated otherwise (\Cref{subsec:distribution_shift}).
\Cref{tab:cameras} gives an overview of the used acquisition hardware and \Cref{tab:used_lights} specifies the used lights for every lighting condition.

\subsubsection*{Can} A large bar light was used to illuminate the scene from above, ensuring homogeneous lighting across the entire field of view. 
The can is positioned in front of a white paper, which helps to create a diffuse and even illumination.
For the acquisition of the test data, two additional spotlights are directed at the can to generate specular highlights on its metallic surface and its label.
\subsubsection*{Fabric} \textit{Fabric} is captured using a flat light that illuminates the fabric piece from above. 
Two additional spotlights are installed for the acquisition of the test data. 
The first is positioned at a low angle relative to the fabric's surface to create an illumination gradient across the image. 
The second spotlight is used to create additional spurious light.
\subsubsection*{Fruit Jelly} Images of \textit{Fruit Jelly} are captured using a setup with both back light illumination by a flat light and front light illumination by a spotlight.
This configuration ensures that light is transmitted through the semi-opaque jelly, making internal defects visible, while also highlighting surface defects on the front. 
The plastic cups are placed on a plate of polymethyl methacrylate on which the lower part of the cup is reflected.
If defects are visible in these reflections, they are labeled as well.
To test the robustness of the methods against spurious lighting conditions, for the acquisition of the test data an additional LED spotlight is mounted to illuminate the object from above.
\subsubsection*{Rice} \textit{Rice} is illuminated using a large flat light, ensuring that no disturbing shadows occur.
Two additional spotlights create illumination gradients for the test data.
\subsubsection*{Sheet Metal} To effectively identify small scratches and holes on metallic surfaces, dark field illumination with a low angle of incidence, which reflects the light towards the camera at surface irregularities, is employed using a ring light. 
As for the other objects, we vary the exposure time of the regular lighting for the acquisition of the test data.
Moreover, we use a second ring light with a steeper angle of incidence and vary the exposure time as well.
The two ring lights, employed individually, result in different illumination patterns, particularly at the outline of the sheet metal.
\subsubsection*{Vial} We acquired the images of the vials using a flat light as back light illumination, which allows us to detect fully or partially transparent contaminants, such as small plastic foil particles.
We further installed three LED spotlights for the acquisition of the test data.
The first one is mounted above the vial to create an inhomogeneous top-down illumination.
The two other spots are positioned to create additional specular highlights on the surface of the vial.
\subsubsection*{Wall Plugs} The \textit{Wall Plugs} bulk goods are illuminated from above using a large flat light. 
For the acquisition of the test data, two LED spotlights are positioned at a low incidence angle to the objects, creating a strong brightness gradient across the scene.
\subsubsection*{Walnuts} Similar to the \textit{Wall Plugs} setup, a flat light is used along with two additional LED spotlights for the acquisition of the test data to create irregular illuminations.

\begin{table}[ht]
\centering
\caption{Cameras and lenses used for image acquisition.}\label{tab:cameras}
\begin{adjustbox}{max width = .98\linewidth}
\setlength\tabcolsep{2 pt}
\begin{tabular}{llll}
 \toprule
Object & Camera  Model & Lens Model (focal length)\\
\midrule
Can & IDS GV-5280CP-C-HQ & Tamron MA111F16VIR (16mm)\\
Fabric & IDS GV-5280CP-C-HQ & Tamron MA111F16VIR (16mm)\\
Fruit Jelly & IDS GV-5280CP-C-HQ & Tamron MA111F16VIR (16mm)\\
Rice & Daheng MER2-503-36U3M &  Moritex ML-U1614MP9 (16mm)\\
Sheet Metal & SVS-Vistek exo834MTLGEC & Tamron MA111F16VIR (16mm)\\
Vial & Daheng MER2-503-36U3M &  Moritex ML-U1614MP9 (16mm)\\
Wall Plugs & Daheng MER2-503-36U3M & Moritex ML-U1614MP9 (16mm)\\
Walnuts & IDS GV-5280CP-C-HQ & Computar M1214-MP2 (12mm)\\
\bottomrule
\end{tabular}
\end{adjustbox}
\end{table}

\begin{table}[h]
\centering
\caption{Lights used for the different lighting conditions contained in the test set. For \textit{Sheet Metal}, both ring lights deploy three distinct exposure times.}
\label{tab:used_lights}
\begin{adjustbox}{max width = .98\linewidth}
\renewcommand{\arraystretch}{1.2}
\begin{tabular}{lllll} 
\toprule
\multicolumn{1}{l}{\multirow{2}{*}{Object}} & \multirow{2}{*}{\begin{tabular}[c]{@{}l@{}}regular, under- and\\overexposed\end{tabular}} & \multicolumn{3}{c}{additional light sources  }  \\ 
\cline{3-5}
\multicolumn{1}{c}{}                        &                                                                                             & shift 1   & shift 2   & shift 3                 \\ 
\hline
Can                      & B                                                                                              & B, S1     & B, S2     & B, S1, S2  \\
Fabric                   & F                                                                                              & F, S1     & F, S2     & F, S1, S2  \\
Fruit Jelly              & F, S1                                                                                          & F, S1, S2 & -         & -          \\
Rice                     & F                                                                                              & F, S1     & F, S2     & F, S1, S2  \\
Sheet Metal              & R1                                                                                             & R2        & R2        & R2         \\
Vial                     & F                                                                                              & F, S1     & F, S2     & F, S3      \\
Wall Plugs               & F                                                                                              & F, S1     & F, S2     & F, S1, S2  \\
Walnuts                  & F                                                                                              & F, S1     & F, S2     & F, S1, S2  \\
\bottomrule
\multicolumn{5}{l}{\scriptsize{B: bar light, F: flat light, R: ring light, S: spotlight}} \\
\end{tabular}
\end{adjustbox}
\end{table}

\section{Benchmark}
\label{sec:benchmark}
\subsection{Evaluated Methods}

As a baseline, we benchmark our dataset against student--teacher-based (EfficientAD \cite{batzner2023_efficientad}, Reverse Distillation \cite{deng2022_rd}, Reverse Distillation Revisited \cite{tien2023_rd_revisited}), memory-bank-based (PatchCore \cite{patchcore}), and normalizing-flow-based (MSFlow \cite{zhou2024_msflow}) methods as well as approaches that imitate anomalous data (SimpleNet \cite{Liu_2023_simplenet}, Dual Subspace Re-Projection \cite{Zavrtanik_2022_DSR}).
For each method, we use the official implementation with the default settings reported for \mad\ in the original publication but without scenario-specific tuning of hyperparameters such as training duration.
If not explicitly stated otherwise, images are resized to $\textrm{256}\times\textrm{256}$ pixels.
In the following, we provide an overview of the implementation and parameterization details of each evaluated method.

\subsubsection*{Reverse Distillation (RD)}
We use the official implementation described in \cite{deng2022_rd}. 
The RD architecture requires the input image size to be a multiple of 32. 
For evaluation settings in which this is not the case, we use the next possible smaller image side length for height and width, respectively. 
In \Cref{tab:performance_on_other_ad_datasets}, the results on \mad\ are taken from \cite{deng2022_rd}.
VisA results are derived with the same parameters as for \mad.

\subsubsection*{Reverse Distillation Revisited (RD++)}
We use the official implementation described in \cite{tien2023_rd_revisited}. 
The implementation is based on the RD framework, which imposes similar constraints on input image size. 
We therefore use the same resizing procedure as for RD\@. 
In \Cref{tab:performance_on_other_ad_datasets}, results on \mad\ are taken from \cite{tien2023_rd_revisited}. 
VisA results are derived with the same parameters as for \mad.

\subsubsection*{EfficientAD}
For EfficientAD, we follow the implementation details that are described in \cite{batzner2023_efficientad}. 
As an architecture variant, we choose EfficientAD-S for the evaluations on \madtwo\ due to its lower latency and comparable performance to the slightly larger EfficientAD-M variant.
Nevertheless, in \Cref{tab:performance_on_other_ad_datasets}, we use EfficientAD-M to show the method's overall best performance on \mad\ and VisA.

\subsubsection*{PatchCore}
We use the official implementation described in \cite{patchcore}.
In our evaluations, we use the default ensembling setting, i.e., WideResNet-101, ResNeXT-101, and DenseNet-201 as backbones, approximate greedy coreset subsampling with a sampling ratio set to 0.01\%, and the final aggregated embedding dimensionality of 384. 
In addition, we disable the center cropping to enable the detection of defects occurring at the image borders.

\subsubsection*{MSFlow}
We use the official implementation described in \cite{zhou2024_msflow}.
However, for deriving the results in \Cref{tab:performance_on_other_ad_datasets}, we use a training and inference image size of 256$\times$256 instead of 512$\times$512 to be comparable with the other methods.
VisA results are derived with the same parameters as for \mad.

\subsubsection*{SimpleNet}
We use the official implementation described in \cite{Liu_2023_simplenet} with a WideResNet50 feature extractor. Following \cite{batzner2023_efficientad}, we do not use a scenario-specific training duration and train for 160 epochs.

\subsubsection*{Dual Subspace Re-Projection (DSR)}
We use the official implementation described in \cite{Zavrtanik_2022_DSR}.
However, we only train DSR for the object \textit{Rice} on the largest image size due to the long training duration.

\subsection{Evaluation Metrics}

\subsubsection{Threshold-Independent Metrics}  
\begin{figure*}[ht]
  \centering
  \includegraphics[width=.98\textwidth]{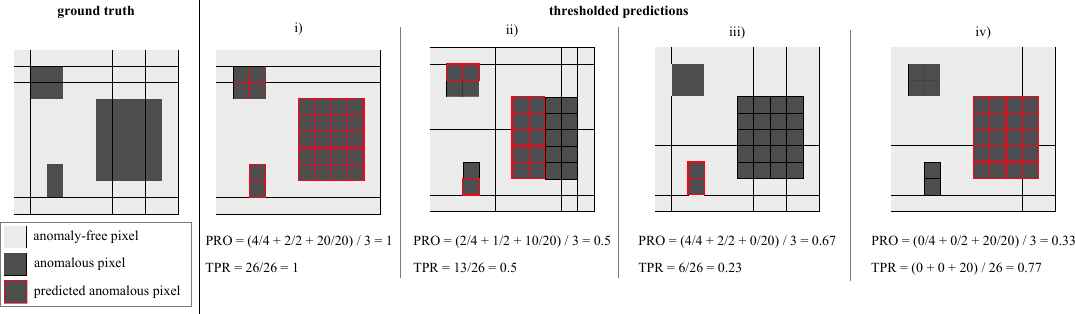}
  \caption{Per region overlap (PRO) vs. true positive rate (TPR) for a given ground truth of anomalous data and different cases of thresholded predictions. PRO considers each anomalous region equally, whereas the TPR is dominated by large defects and overestimates defect localization quality.}
  \label{fig:auc_pro_vs_roc}
\end{figure*}

To assess the localization quality, we employ the area under the per-region-overlap (PRO) curve on the pixel level \cite{bergmann2021_mvtec_ad_ijcv} (\aupro), which integrates the PRO over the false positive rate (FPR) for varying thresholds.
The per-region-overlap is defined as 
\begin{equation}
    \mathrm{PRO}=\frac{1}{k} \sum_{i=1}^{\left|D_{\text {test }}\right|} \sum_{j=1}^{k_i} \frac{\left|P_{\mathrm{ano}} \cap C_{i, j}\right|}{\left|C_{i, j}\right|}
\end{equation}
where $C_{i, j}$ represents the set of pixels classified as anomalous for a connected component $j$ in the ground truth image $i$ and $ P_{\mathrm{ano}}$ is the set of pixels predicted as anomalous for a given threshold $t$. 
In contrast to the area under the receiver operator curve (AU-ROC) that can also be used to evaluate anomaly localization performance on the pixel level and measures the true positive rate (TPR) against the false positive rate, \aupro\ considers each anomalous region equally.
In case of multiple anomalous regions the TPR is dominated by large anomalies and overestimates localization quality, because the majority of the defects might not be detected at all.
In contrast, the PRO is not corrupted by differently sized anomalies (\Cref{fig:auc_pro_vs_roc}).
Since industrial applications require the detection of every single defect regardless of its size, we choose \aupro\ over \mbox{AU-ROC}.

\begin{figure}[ht]
  \centering
  \includegraphics[width = \columnwidth]{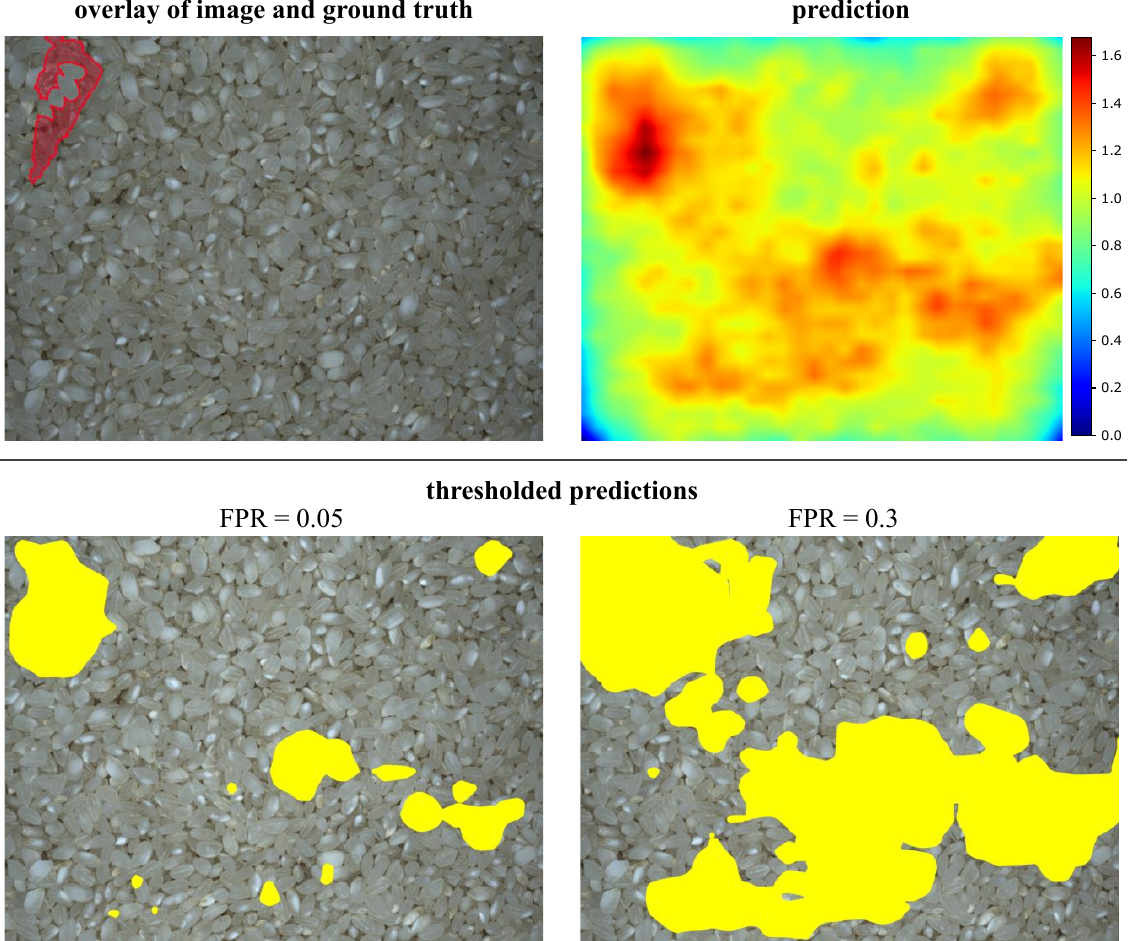}
  \caption{Reducing the integration limit of \aupro\ to foster more meaningful anomaly maps. An exemplary anomaly map of MSFlow on the object \textit{Rice} for an image size of $\textsf{256}\times\textsf{256}$ is shown and thresholded to obtain the desired false positive rate (FPR). The common integration limit $\textsf{FPR} = \textsf{0.3}$ allows segmented defects that are drastically too large.}
  \label{fig:reduced_aupro_integration_limit}
\end{figure}

Typically, the PRO curve is integrated up to a false positive rate (FPR) of 0.3 (\auprothirty).
However, because of the utmost importance of an accurate anomaly segmentation to ensure the high quality standards in modern industrial production systems, we decrease the tolerance for false positive pixels and establish an integration limit for the PRO curve of 0.05 for \madtwo\ (\auprofive).
\Cref{fig:reduced_aupro_integration_limit} illustrates the effect of a reduced integration limit.
Maps created with a FPR of 30\% do not conform to human expectations of what a meaningful anomaly map should look like at all.
Besides, setting a tighter integration limit is especially important for small defects in large images as contained in \madtwo.
For example, for a defect size of 10 pixels in a 5MP image, even a FPR of 5\% as integration limit results in a wrongly segmented area that is more than 25,000 times larger than the defect itself.
Nevertheless, \Cref{tab:performance_on_other_ad_datasets} refers to \auprothirty\ to highlight the difficulty of our dataset even with respect to the more tolerant evaluation setting with detailed values provided in the supplemental material.

\subsubsection{Threshold-Dependent Metrics} 
Threshold-independent metrics such as \aupro\ are particularly valuable to evaluate the general quality of anomaly maps.
However, in industrial practice, the primary goal of an anomaly detection system is to determine whether the examined part is defective or defect-free based on an informed and explainable decision.
Thus, it is essential to apply thresholds to separate these two classes on both the pixel and image level.
For this reason, we obtain binary segmentation maps by thresholding the anomaly images and compare them to the ground truth data by computing the harmonic mean of precision and recall ($F_{1}$) on the pixel level.
Since images that contain defects are unavailable, as a baseline the segmentation threshold is set to the average of the per-pixel anomaly values of the defect-free validation images plus three times their standard deviation.
From a practical point of view, it is reasonable to expect that the global classification as \textit{good} or \textit{reject} is derived from the detection of an anomalous region.
Hence, the segmentation threshold simultaneously serves as the classification threshold, where the maximum pixel value is considered as the image anomaly score.
Again, $F_{1}$ is used to measure image-level classification performance after thresholding.

\subsubsection{Runtime and Memory Consumption}
\label{subsubsec:runtime_and_memory_consumption}
The deployment of anomaly detection methods in practice requires a thorough consideration of the available hardware since compute power and memory might be limited.
Consequently, runtime and memory footprint quickly become critical constraints.
Therefore, we measure the inference time for a single image as well as peak memory usage of all methods on an NVIDIA RTX 2080Ti.

Here, we generally follow the procedure described in \cite{batzner2023_efficientad} and also provide code snippets for the measurements.
However, since our focus lies on showing the effects of using larger image sizes, we did not optimize model architectures for inference efficiency.
In particular, we measure the inference runtime as the mean of 1000 forward passes after a warm up of the same duration with a batch size of~1 using float32.
Timing begins before transferring the dataset sample to the GPU and stops after the anomaly image is available on the CPU using Python's {\fontfamily{qcr}\selectfont time} module.
Memory is measured with the official PyTorch functionality ({\fontfamily{qcr}\selectfont torch.cuda.memory stats()[’reserved bytes.all.peak’]}).
Note that for PatchCore, the inference runtime depends on the size of the memory bank that is defined by the number of training images. Thus, we combined \testpublic\ and \testprivate\ and, therefore, computed the average over inferring approximately 200 samples after warm up.

\subsection{Results}

\subsubsection{Threshold-Independent Metrics}
\Cref{tab:results_aucpro_005} provides the anomaly localization results  \auprofive\ of the evaluated methods on \madtwo.
EfficientAD achieves the best overall performance in the regular setting of unchanged lighting conditions, with a mean performance of  30.8\% \auprofive\ on \testprivate, followed by PatchCore, RD++, RD, MSFlow, SimpleNet, and DSR.
The results show a large room for improvement in localization quality in general. The object \textit{Can} is especially challenging for the investigated state-of-the-art models.
\Cref{tab:results_aucpro_005} also indicates that certain methods like RD are more robust against distribution shifts induced by lighting condition changes.
RD performs only 1.4 percentage points worse on \testprivatemixed, whereas the \auprofive\ of MSFlow drops by 12.4 percentage points from 24.3\% to 11.9\%.
Observing such variations in robustness, which previously could not be investigated thoroughly, not only demonstrates the contribution of \madtwo\ but also underlines the importance of considering the effects of changing environmental conditions in future method development.

\begin{table*}[ht]
\centering
\caption{Segmentation \auprofive\ results (in \%) for \testprivate\ / \testprivatemixed\ set. All methods achieve less than 31\% segmentation performance on the \testprivate\ of \madtwo. In addition, the performance gaps between \testprivate\ and \testprivatemixed\ highlight the different robustness of methods to changes in lighting conditions.} \label{tab:results_aucpro_005}
\renewcommand{\arraystretch}{1.1}
\begin{adjustbox}{max width = .98\textwidth}
\begin{tabular}{lccccccc}
 \toprule
Object & PatchCore  & RD  & RD++ & EfficientAD & MSFlow & SimpleNet & DSR \\
\midrule
Can & 4.7 / 4.6 & 7.0 / 7.5 & 7.7 / 7.0 & 9.6 / 1.3 & 6.7 / 0.8 & 8.4 / 1.9 & 13.9 / 3.5 \\
Fabric & 11.0 / 12.0 & 3.5 / 3.6 & 4.8 / 5.3 & 22.2 / 13.0 & 14.1 / 14.3 & 6.6 / 7.3 & 6.8 / 5.3 \\
Fruit Jelly & 46.7 / 46.7 & 48.2 / 48.2 & 54.4 / 54.4 & 50.5 / 47.6 & 49.4 / 38.3 & 39.8 / 38.5 & 36.0 / 34.2 \\
Rice & 25.6 / 18.5 & 11.2 / 11.4 & 12.2 / 10.2 & 27.6 / 4.3 & 21.5 / 12.2 & 8.7 / 4.2 & 7.8 / 8.3 \\
Sheet Metal & 15.2 / 13.0 & 9.5 / 8.8 & 9.3 / 8.9 & 11.8 / 5.3 & 11.6 / 7.7 & 12.0 / 9.1 & 18.0 / 16.1 \\
Vial & 62.2 / 59.1 & 62.1 / 60.3 & 63.0 / 57.0 & 55.6 / 47.7 & 38.0 / 4.0 & 47.8 / 23.3 & 50.0 / 48.1 \\
Wall Plugs & 12.8 / 9.9 & 19.7 / 12.2 & 15.4 / 10.1 & 20.3 / 1.2 & 12.6 / 0.2 & 5.7 / 1.9 & 3.9 / 6.5 \\
Walnuts & 51.8 / 44.5 & 50.2 / 48.3 & 49.7 / 48.9 & 48.8 / 33.0 & 40.4 / 18.1 & 40.0 / 23.4 & 25.7 / 17.1 \\
\midrule
Mean & 28.8 / 26.0 & 26.4 / 25.0 & 27.1 / 25.2 & 30.8 / 19.2 & 24.3 / 11.9 & 21.1 / 13.7 & 20.3 / 17.4 \\
\bottomrule
\end{tabular}
\end{adjustbox}
\vspace{6.5mm}
\end{table*}

\vspace{-1mm}
\subsubsection{Threshold-Dependent Metrics}
In \Cref{tab:results_seg_f1}, we report the $F_{1}$ score on the pixel level for the thresholded anomaly maps.
Here, MSFlow shows the best overall performance in the regular setting with a $F_{1}$ score of 21.8\% on \testprivate.
Apart from the challenges posed by changing lighting conditions, our analysis reveals that performance measured after applying a threshold does not necessarily correlate with performance measured by threshold-independent metrics.
For instance, PatchCore performs significantly worse than all other methods in terms of $F_{1}$ score, yet achieves similar results in terms of \aupro\@.
Consequently, the sole evaluation of anomaly map predictions based on threshold-independent metrics is insufficient to assess their real-world applicability.
Therefore, the development of new anomaly detection methods and the determination of a suitable threshold should always be addressed together.

\begin{table*}[ht]
\centering
\caption{Segmentation $F_{1}$ score (in \%) on binarized images for \testprivate\ / \testprivatemixed\ set. The segmentation threshold that decides whether a pixel is categorized as \textit{normal} or \textit{anomalous} is determined on the validation set, i.e., using defect-free images only.} \label{tab:results_seg_f1}
\renewcommand{\arraystretch}{1.1}
\begin{adjustbox}{max width = .98\linewidth}
\begin{tabular}{lccccccc}
 \toprule
Object & PatchCore  & RD  & RD++ & EfficientAD & MSFlow & SimpleNet & DSR \\
\midrule
Can & 0.3 / 0.1 & 0.1 / 0.1 & 0.1 / 0.1 & 0.8 / 0.1 & 5.0 / 0.1 & 0.6 / 0.1 & 0.4 / 0.1 \\
Fabric & 11.5 / 9.8 & 2.6 / 2.2 & 2.9 / 2.3 & 7.6 / 1.0 & 22.0 / 4.1 & 21.6 / 10.2 & 7.9 / 5.0 \\
Fruit Jelly & 8.7 / 8.2 & 22.5 / 22.7 & 26.9 / 26.7 & 20.8 / 18.2 & 47.6 / 38.1 & 25.1 / 23.0 & 17.9 / 17.2 \\
Rice & 3.8 / 4.2 & 7.0 / 3.9 & 9.5 / 2.9 & 15.0 / 0.5 & 19.1 / 1.8 & 11.6 / 1.0 & 1.5 / 1.4 \\
Sheet Metal & 1.8 / 1.1 & 41.3 / 39.2 & 40.9 / 37.7 & 9.3 / 3.8 & 13.0 / 7.6 & 14.6 / 2.8 & 13.9 / 14.4 \\
Vial & 2.3 / 2.2 & 28.0 / 28.3 & 28.2 / 22.8 & 30.5 / 26.5 & 23.3 / 6.2 & 31.9 / 17.5 & 28.2 / 27.9 \\
Wall Plugs & 0.0 / 0.0 & 1.9 / 0.8 & 1.3 / 0.9 & 4.4 / 0.3 & 0.1 / 0.2 & 1.0 / 0.3 & 0.4 / 0.4 \\
Walnuts & 1.2 / 1.3 & 41.2 / 36.7 & 44.1 / 40.5 & 34.6 / 13.3 & 44.5 / 14.3 & 35.2 / 14.3 & 17.0 / 9.6 \\
\midrule
Mean & 3.7 / 3.4 & 18.1 / 16.7 & 19.2 / 16.7 & 15.4 / 8.0 & 21.8 / 9.0 & 17.7 / 8.7 & 10.9 / 9.5 \\

\bottomrule
\end{tabular}
\end{adjustbox}
\vspace{6.5mm}
\end{table*}

\vspace{-1mm}
\subsubsection{Reaching Higher Performance at Higher Cost}
Our results indicate that current state-of-the-art methods are challenged by the anomaly detection scenarios contained in \madtwo.
However, since, among several other criteria, our dataset design focused on small and low-contrast defects, we hypothesized that certain anomalies simply disappear during the preprocessing to scale the images to the current standard setting of an image size of $\textrm{256} \times \textrm{256}$.
Hence, we conducted experiments with an increased input size up to half the original image height and width.

\begin{figure}[ht]
\centering
\includegraphics[width=.9\columnwidth]{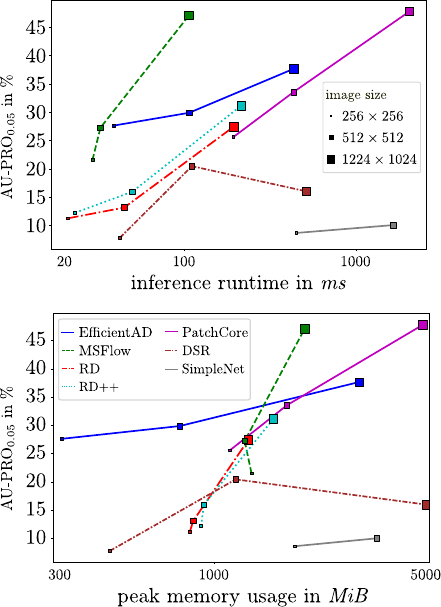}
\caption{Scaling up the input: Inference times and memory usage quickly exceed real-time requirements of many industrial production cycles (logarithmic x-axis). 
Inference times and peak GPU memory usage are measured on a NVIDIA RTX 2080 Ti GPU for the \madtwo\ object \textit{Rice}.}
\label{fig:high_perf_at_higher_cost}
\end{figure}

As \Cref{fig:high_perf_at_higher_cost} illustrates for the object \textit{Rice}, processing higher resolutions can boost performance significantly.
For example, the \auprofive\ of PatchCore and MSFlow approximately double for the largest image size.
However, at the same time, the inference runtime and memory consumption drastically increase by more than an order of magnitude. 
For instance, PatchCore achieves the highest \auprofive\ but inference takes 2~seconds per image. SimpleNet, on the other hand, exceeds the available memory for the largest image size.
In practical industrial applications, which almost always operate with limited compute or memory resources or require very short runtimes, this might prohibit deploying such models at all.  
Notably, MSFlow scales much more efficiently while achieving similar localization performance. 
To further establish runtime and memory usage as important performance indicators of anomaly detection methods, we provide code snippets to record runtime and memory usage in a more standardized way and require participants of our benchmark to disclose the used image size as a proxy for these characteristics when entering the \madtwo\ leader board.

\section{Evaluation Server for Standardized and Fair Performance Assessment}
\label{sec:benchmark_server}

As outlined in \Cref{subsec:dataset_description}, we provide an evaluation server that holds the pixel-precise ground truth for the two private test sets \testprivate\ and \testprivatemixed. 
The primary objective of this server is to ensure a fair and standardized comparison of anomaly detection methods on \madtwo. 
Since only image data is available for these test sets, the evaluation server emphasizes the unsupervised anomaly detection setting, where the nature of defects is unknown until the anomaly detection systems are deployed. By maintaining a private test set, we ensure the integrity of the evaluation process by preventing overfitting and excessive tuning on the test data.
In practice, accurate anomaly localization is crucial for establishing trust in a detection system's decision about whether an object is defective or good.
Hence, we focus on the evaluation of segmentation performance. 
In addition to the threshold-independent \aupro\ evaluation, we encourage research on robust methods to derive thresholds for the anomaly map, which ultimately determines the classification of an object. 
Therefore, we provide the possibility to submit a segmentation threshold along with the continuous anomaly maps for the test sets.
\\
\\
The procedure to evaluate model performance on \testprivate\ or \testprivatemixed\ and to enter the \madtwo\ leader board is as follows:
\begin{enumerate}
    \item Train the model on anomaly-free training data.
    \item Use defect-free validation images to derive hyperparameters such as training duration or thresholds for binary segmentation. Optionally, use \testpublic\ for initial performance estimation.
    \item Run the inference on the \testprivate\ and/or \testprivatemixed\ image data and store resulting anomaly maps as \textit{TIFF} files in the required directory structure:\\
    {\fontfamily{qcr}\selectfont 
    /mvtec\_ad\_2/\textit{\{object\_name\}}/\\
    \{\textit{private,private\_mixed\}}/\\
    anomaly\_images/test/\{\textit{good,bad}\}/\\
    \{\textit{image\_name}\}.tiff
    }
    \item Apply the local data structure checks. If successful, the directory is compressed automatically into a zip folder.  You can download the necessary code from the dataset website\footnote[1]{\url{https://www.mvtec.com/company/research/datasets/mvtec-ad-2}}
    \item Upload the zipped data.
    \item Receive the performance results.
\end{enumerate}
\noindent
The evaluation server is accessible at \url{https://benchmark.mvtec.com/}.
Further instructions can be found in the supplemental material or on the dataset website\footnotemark[1].

\section{Conclusion}
\label{sec:conclusion}
We introduce \madtwo, a new dataset for unsupervised anomaly detection. It extends existing datasets by previously unrepresented advanced scenarios, including transparent and overlapping objects, dark-field and back light illumination, objects with high variance in the normal data, extremely small defects compared to the image size, and anomalies occurring throughout the entire image area.
Focusing on the localization performance, the results in our benchmark of seven state-of-the-art methods suggest a large room for improvements.
Anomaly localization measured in \auprothirty\ remains below 60\%, and, when reducing the tolerance for false positive segmentation results by applying \auprofive, even below approximately 30\%.
Although performance on \madtwo\ can generally be increased by using larger image sizes, we highlighted the necessity of efficiently scaling up current models with respect to runtime and memory consumption.
Additionally, for the first time, non-synthetic test data with lighting condition changes enables an investigation of the robustness of methods against real-world distribution shifts.
Moreover, we host an evaluation server that not only considers threshold-independent but also threshold-dependent metrics. This contributes towards a fairer and more standardized assessment of anomaly detection methods within the research community.

\subsubsection*{Limitations and Future Work}
Although we benchmark \madtwo\ on seven distinct state-of-the-art models, we intend to report results for even more methods on the evaluation server.
We also believe that using a more sophisticated method for threshold estimation than the one used in our experiments might improve the correlation between threshold-independent and threshold-dependent metrics.
What is more, we see a large room for innovative method design eliminating the need for threshold selection in general.
For methods displaying large differences in performance between regular exposure (\testprivate) and changed lighting conditions (\testprivatemixed), it would be interesting to investigate how extensive data augmentation with standard techniques like brightness variation could close this gap, or if advanced augmentation techniques are required to offer a possible solution.
Additionally, it would be worth exploring whether augmentation adversely affects performance on the regular test set.
For large input images, efficient processing strategies such as tiling or multi-scale approaches need to be investigated further.
Finally, to measure efficiency in terms of inference runtime and memory consumption, we propose the creation of publicly available benchmark servers as a standardized hardware environment.
These should potentially include embedded devices such as the NVIDIA Jetson Nano\footnote{\url{https://developer.nvidia.com/embedded/jetson-nano}} with a memory of just 4~GB or the Hailo-8 AI Accelerator\footnote{\url{https://hailo.ai/products/ai-accelerators/hailo-8-ai-accelerator}} that operates with int8 quantization to reflect common industrial hardware constraints adequately. 

\newpage
\bibliographystyle{IEEEtran}
\bibliography{egbib}

\begin{thebibliography}{10}
\providecommand{\url}[1]{#1}
\csname url@samestyle\endcsname
\providecommand{\newblock}{\relax}
\providecommand{\bibinfo}[2]{#2}
\providecommand{\BIBentrySTDinterwordspacing}{\spaceskip=0pt\relax}
\providecommand{\BIBentryALTinterwordstretchfactor}{4}
\providecommand{\BIBentryALTinterwordspacing}{\spaceskip=\fontdimen2\font plus
\BIBentryALTinterwordstretchfactor\fontdimen3\font minus \fontdimen4\font\relax}
\providecommand{\BIBforeignlanguage}[2]{{%
\expandafter\ifx\csname l@#1\endcsname\relax
\typeout{** WARNING: IEEEtran.bst: No hyphenation pattern has been}%
\typeout{** loaded for the language `#1'. Using the pattern for}%
\typeout{** the default language instead.}%
\else
\language=\csname l@#1\endcsname
\fi
#2}}
\providecommand{\BIBdecl}{\relax}
\BIBdecl

\bibitem{blum2019_fishyscapes_dataset}
H.~Blum, P.-E. Sarlin, J.~Nieto, R.~Siegwart, and C.~Cadena, ``Fishyscapes: A benchmark for safe semantic segmentation in autonomous driving,'' in \emph{IEEE International Conference on Computer Vision Workshops (ICCVW)}, 2019, pp. 2403--2412.

\bibitem{hendrycks2019_benchmark_anomaly_segmentation_autodriving}
D.~Hendrycks, S.~Basart, M.~Mazeika, M.~Mostajabi, J.~Steinhardt, and D.~Song, ``A benchmark for anomaly segmentation,'' \emph{arXiv preprint arXiv:1911.11132v1}, 2019.

\bibitem{seebock2019_uncertainty_for_ad}
P.~Seeböck, J.~I. Orlando, T.~Schlegl, S.~M. Waldstein, H.~Bogunović, S.~Klimscha, G.~Langs, and U.~Schmidt-Erfurth, ``Exploiting epistemic uncertainty of anatomy segmentation for anomaly detection in retinal {OCT},'' \emph{IEEE Transactions on Medical Imaging}, vol.~39, no.~1, pp. 87--98, 2020.

\bibitem{menze2015_brats_dataset}
B.~H. Menze, A.~Jakab, S.~Bauer, J.~Kalpathy-Cramer, K.~Farahani, J.~Kirby \emph{et~al.}, ``The multimodal brain tumor image segmentation benchmark ({BRATS}),'' \emph{IEEE Transactions on Medical Imaging}, vol.~34, no.~10, pp. 1993--2024, 2015.

\bibitem{nazare2018_pretrained_cnns_for_ad}
T.~S. Nazare, R.~F. de~Mello, and M.~A. Ponti, ``Are pre-trained {CNNs} good feature extractors for anomaly detection in surveillance videos?'' \emph{arXiv preprint arXiv:1811.08495}, 2018.

\bibitem{li2013_ucsd_video_ad_dataset}
W.-X. Li, V.~Mahadevan, and N.~Vasconcelos, ``Anomaly detection and localization in crowded scenes,'' \emph{IEEE Transactions on Pattern Analysis and Machine Intelligence}, vol.~36, no.~1, pp. 18--32, 2013.

\bibitem{bergmann2019_mvtec_ad_cvpr}
P.~Bergmann, M.~Fauser, D.~Sattlegger, and C.~Steger, ``{MVTec AD} --- {A} comprehensive real-world dataset for unsupervised anomaly detection,'' in \emph{IEEE Conference on Computer Vision and Pattern Recognition (CVPR)}, 2019, pp. 9584--9592.

\bibitem{Zou_2022_VisA}
Y.~Zou, J.~Jeong, L.~Pemula, D.~Zhang, and O.~Dabeer, ``Spot-the-difference self-supervised pre-training for anomaly detection and segmentation,'' in \emph{Computer Vision -- ECCV 2022}.\hskip 1em plus 0.5em minus 0.4em\relax Cham: Springer Nature Switzerland, 2022, pp. 392--408.

\bibitem{bergmann2021_mvtec_ad_ijcv}
P.~Bergmann, K.~Batzner, M.~Fauser, D.~Sattlegger, and C.~Steger, ``{The MVTec Anomaly Detection Dataset: A Comprehensive Real-World Dataset for Unsupervised Anomaly Detection},'' \emph{International Journal of Computer Vision}, vol. 129, no.~4, pp. 1038--1059, 2021.

\bibitem{bergmann2021_mvtec_loco_ijcv}
------, ``Beyond dents and scratches: Logical constraints in unsupervised anomaly detection and localization,'' \emph{International Journal of Computer Vision}, vol. 130, no.~4, p. 947–969, 2022.

\bibitem{bergmann2022_mvtec_3d_ad}
P.~Bergmann, X.~Jin, D.~Sattlegger, and C.~Steger, ``The {MVTec 3D-AD} dataset for unsupervised {3D} anomaly detection and localization,'' in \emph{Proceedings of the 17th International Joint Conference on Computer Vision, Imaging and Computer Graphics Theory and Applications --- Volume 5: VISAPP}.\hskip 1em plus 0.5em minus 0.4em\relax SciTePress, 2022, pp. 202--213.

\bibitem{bonfiglioli_eyecandies_2022}
L.~Bonfiglioli, M.~Toschi, D.~Silvestri, N.~Fioraio, and D.~De~Gregorio, ``The {Eyecandies} dataset for unsupervised multimodal anomaly detection and localization,'' in \emph{Proceedings of the 16th Asian Conference on Computer Vision (ACCV 2022)}, 2022.

\bibitem{cpr}
H.~Li, J.~Hu, B.~Li, H.~Chen, Y.~Zheng, and C.~Shen, ``Target before shooting: Accurate anomaly detection and localization under one millisecond via cascade patch retrieval,'' \emph{IEEE Transactions on Image Processing}, vol.~33, pp. 5606--5621, 2024.

\bibitem{batzner2023_efficientad}
K.~Batzner, L.~Heckler, and R.~K\"onig, ``{EfficientAD}: Accurate visual anomaly detection at millisecond-level latencies,'' in \emph{Proceedings of the IEEE/CVF Winter Conference on Applications of Computer Vision (WACV)}, 2024, pp. 128--138.

\bibitem{zhang2023_diffusionad}
H.~Zhang, Z.~Wang, Z.~Wu, and Y.-G. Jiang, ``{DiffusionAD}: Norm-guided one-step denoising diffusion for anomaly detection,'' \emph{arXiv preprint arXiv:2303.08730v3}, 2023.

\bibitem{wang_real-iad_2024}
C.~Wang, W.~Zhu, B.-B. Gao, Z.~Gan, J.~Zhang, Z.~Gu, S.~Qian, M.~Chen, and L.~Ma, ``{Real-IAD}: A real-world multi-view dataset for benchmarking versatile industrial anomaly detection,'' \emph{arXiv preprint arXiv:2403.12580}, 2024.

\bibitem{deng2022_rd}
H.~Deng and X.~Li, ``Anomaly detection via reverse distillation from one-class embedding,'' in \emph{IEEE Conference on Computer Vision and Pattern Recognition (CVPR)}, June 2022, pp. 9737--9746.

\bibitem{zhou2023_pad}
Q.~Zhou, W.~Li, L.~Jiang, G.~Wang, G.~Zhou, S.~Zhang, and H.~Zhao, ``{PAD}: A dataset and benchmark for pose-agnostic anomaly detection,'' in \emph{Thirty-seventh Conference on Neural Information Processing Systems Datasets and Benchmarks Track}, 2023.

\bibitem{bergmann2018_ssim_ae}
P.~Bergmann, S.~Löwe, M.~Fauser, D.~Sattlegger, and C.~Steger, ``Improving unsupervised defect segmentation by applying structural similarity to autoencoders,'' in \emph{Proceedings of the 14th International Joint Conference on Computer Vision, Imaging and Computer Graphics Theory and Applications --- Volume 5: VISAPP}.\hskip 1em plus 0.5em minus 0.4em\relax SciTePress, 2019, pp. 372--380.

\bibitem{Zavrtanik_DRAEM}
V.~Zavrtanik, M.~Kristan, and D.~Sko\v{c}aj, ``{DRAEM} --- {A} discriminatively trained reconstruction embedding for surface anomaly detection,'' in \emph{Proceedings of the IEEE/CVF International Conference on Computer Vision (ICCV)}, October 2021, pp. 8330--8339.

\bibitem{zhang2023_diffad}
X.~Zhang, N.~Li, J.~Li, T.~Dai, Y.~Jiang, and S.-T. Xia, ``Unsupervised surface anomaly detection with diffusion probabilistic model,'' in \emph{IEEE Conference on Computer Vision and Pattern Recognition (CVPR)}, 2023, pp. 6782--6791.

\bibitem{patchcore}
K.~Roth, L.~Pemula, J.~Zepeda, B.~Sch{\"{o}}lkopf, T.~Brox, and P.~V. Gehler, ``Towards total recall in industrial anomaly detection,'' in \emph{IEEE Conference on Computer Vision and Pattern Recognition (CVPR)}.\hskip 1em plus 0.5em minus 0.4em\relax {IEEE}, 2022, pp. 14\,298--14\,308.

\bibitem{cohen2020_spade}
N.~Cohen and Y.~Hoshen, ``Sub-image anomaly detection with deep pyramid correspondences,'' \emph{arXiv preprint arXiv:2005.02357v1}, 2020.

\bibitem{padim}
T.~Defard, A.~Setkov, A.~Loesch, and R.~Audigier, ``{PaDiM}: {A} patch distribution modeling framework for anomaly detection and localization,'' in \emph{Pattern Recognition. {ICPR} International Workshops and Challenges 2021, Proceedings, Part {IV}}, ser. Lecture Notes in Computer Science, vol. 12664.\hskip 1em plus 0.5em minus 0.4em\relax Springer, 2020, pp. 475--489.

\bibitem{Gudovskiy_2022_CFlow}
D.~Gudovskiy, S.~Ishizaka, and K.~Kozuka, ``{CFLOW-AD}: Real-time unsupervised anomaly detection with localization via conditional normalizing flows,'' in \emph{2022 IEEE/CVF Winter Conference on Applications of Computer Vision (WACV)}, 2022, pp. 1819--1828.

\bibitem{csflow}
M.~Rudolph, T.~Wehrbein, B.~Rosenhahn, and B.~Wandt, ``Fully convolutional cross-scale-flows for image-based defect detection,'' in \emph{{IEEE/CVF} Winter Conference on Applications of Computer Vision (WACV)}.\hskip 1em plus 0.5em minus 0.4em\relax {IEEE}, 2022, pp. 1829--1838.

\bibitem{fastflow}
J.~Yu, Y.~Zheng, X.~Wang, W.~Li, Y.~Wu, R.~Zhao, and L.~Wu, ``{FastFlow}: Unsupervised anomaly detection and localization via {2D} normalizing flows,'' \emph{arXiv preprint arXiv:2111.07677v2}, 2021.

\bibitem{zhou2024_msflow}
Y.~Zhou, X.~Xu, J.~Song, F.~Shen, and H.~T. Shen, ``{MSFlow}: Multiscale flow-based framework for unsupervised anomaly detection,'' \emph{IEEE Transactions on Neural Networks and Learning Systems}, pp. 1--14, 2024.

\bibitem{bergmann2020_uninformed_cvpr}
P.~Bergmann, M.~Fauser, D.~Sattlegger, and C.~Steger, ``Uninformed students: Student--teacher anomaly detection with discriminative latent embeddings,'' in \emph{IEEE Conference on Computer Vision and Pattern Recognition (CVPR)}, 2020, pp. 4182--4191.

\bibitem{asymST}
M.~Rudolph, T.~Wehrbein, B.~Rosenhahn, and B.~Wandt, ``Asymmetric student-teacher networks for industrial anomaly detection,'' in \emph{{IEEE/CVF} Winter Conference on Applications of Computer Vision (WACV)}.\hskip 1em plus 0.5em minus 0.4em\relax {IEEE}, 2023, pp. 2591--2601.

\bibitem{tien2023_rd_revisited}
T.~D. Tien, A.~T. Nguyen, N.~H. Tran, T.~D. Huy, S.~T. Duong, C.~D.~T. Nguyen, and S.~Q.~H. Truong, ``Revisiting reverse distillation for anomaly detection,'' in \emph{IEEE Conference on Computer Vision and Pattern Recognition (CVPR)}, June 2023, pp. 24\,511--24\,520.

\bibitem{jeong2023_winclip}
J.~Jeong, Y.~Zou, T.~Kim, D.~Zhang, A.~Ravichandran, and O.~Dabeer, ``Winclip: Zero-/few-shot anomaly classification and segmentation,'' in \emph{IEEE Conference on Computer Vision and Pattern Recognition (CVPR)}, 2023, pp. 19\,606--19\,616.

\bibitem{gu2024_anomalygpt}
Z.~Gu, B.~Zhu, G.~Zhu, Y.~Chen, M.~Tang, and J.~Wang, ``{AnomalyGPT}: Detecting industrial anomalies using large vision-language models,'' \emph{Proceedings of the AAAI Conference on Artificial Intelligence}, vol.~38, no.~3, pp. 1932--1940, 2024.

\bibitem{Schlueter_2022_nsa}
H.~M. Schl{\"u}ter, J.~Tan, B.~Hou, and B.~Kainz, ``Natural synthetic anomalies for self-supervised anomaly detection and localization,'' in \emph{Computer Vision -- ECCV 2022}, S.~Avidan, G.~Brostow, M.~Ciss{\'e}, G.~M. Farinella, and T.~Hassner, Eds.\hskip 1em plus 0.5em minus 0.4em\relax Cham: Springer Nature Switzerland, 2022, pp. 474--489.

\bibitem{Zavrtanik_2022_DSR}
V.~Zavrtanik, M.~Kristan, and D.~Sko{\v{c}}aj, ``{DSR} --- {A} dual subspace re-projection network for surface anomaly detection,'' in \emph{Computer Vision -- ECCV 2022}, S.~Avidan, G.~Brostow, M.~Ciss{\'e}, G.~M. Farinella, and T.~Hassner, Eds.\hskip 1em plus 0.5em minus 0.4em\relax Cham: Springer Nature Switzerland, 2022, pp. 539--554.

\bibitem{Liu_2023_simplenet}
Z.~Liu, Y.~Zhou, Y.~Xu, and Z.~Wang, ``Simplenet: A simple network for image anomaly detection and localization,'' in \emph{Proceedings of the IEEE/CVF Conference on Computer Vision and Pattern Recognition (CVPR)}, June 2023, pp. 20\,402--20\,411.

\bibitem{steger:18}
C.~Steger, M.~Ulrich, and C.~Wiedemann, \emph{Machine Vision Algorithms and Applications}, 2nd~ed.\hskip 1em plus 0.5em minus 0.4em\relax Weinheim: Wiley-VCH, 2018.

\end{thebibliography}

\begin{IEEEbiography}[{\includegraphics[width=1in,height=1.25in,clip,keepaspectratio]{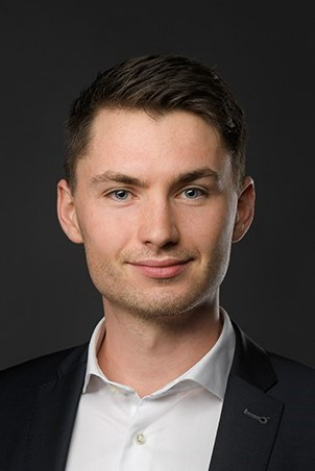}}]
{Lars Heckler-Kram}
studied mechatronics and robotics at the Technical University of Munich (TUM) and received his Master's degree in 2022. He is currently working toward the PhD degree with MVTec Software GmbH and the TUM School of Computation, Information and Technology. His research interests focus on industrial visual anomaly detection.
\end{IEEEbiography}

\begin{IEEEbiography}[{\includegraphics
[width=1in,height=1.25in,keepaspectratio]{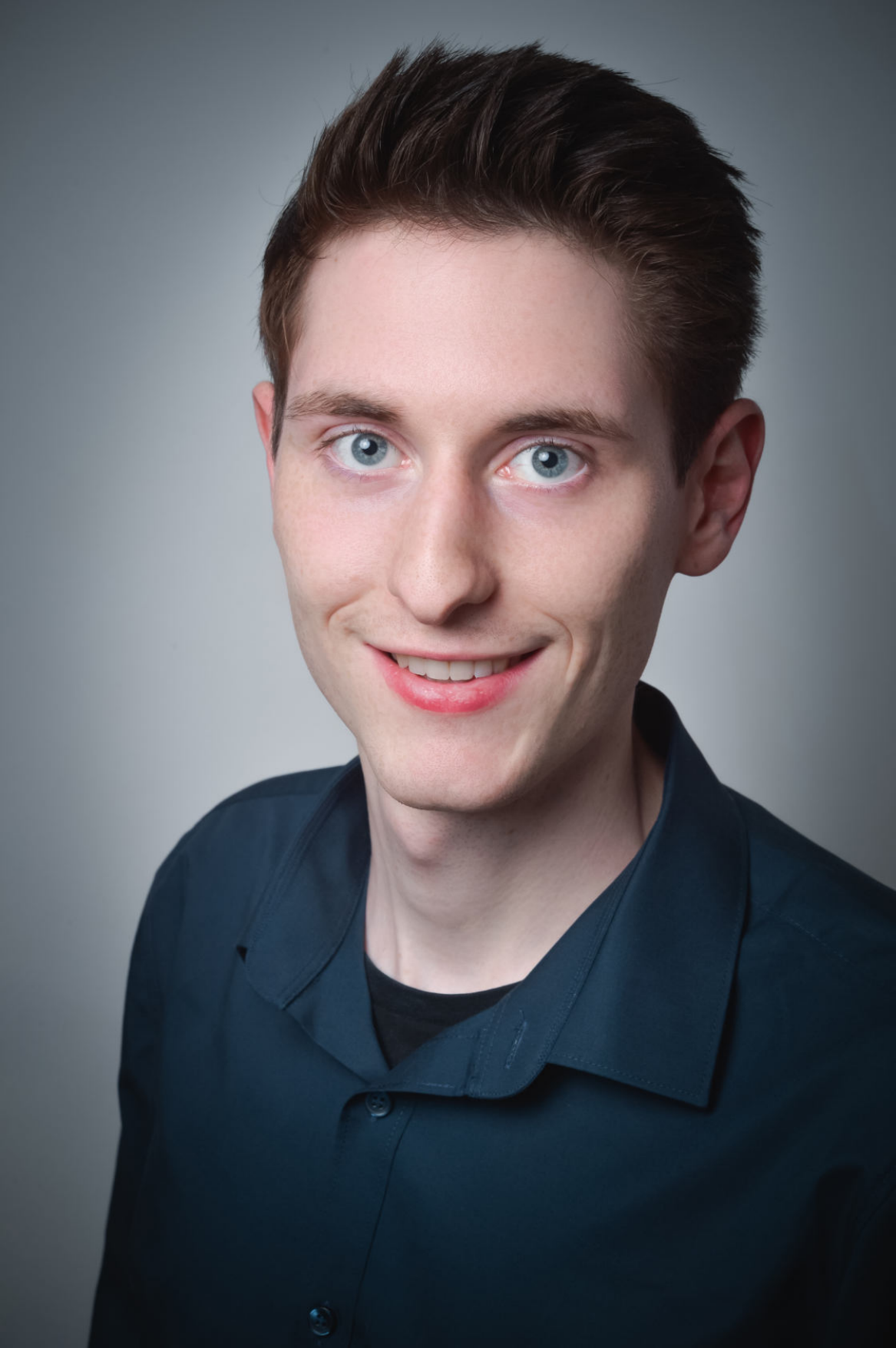}}]
{Jan-Hendrik Neudeck}
received his Master’s degree in Computer Science with a specialization in computer vision and deep learning from the Technical University of Munich (TUM) in 2020. Since then, he has been working as a research engineer at MVTec Software GmbH.
\end{IEEEbiography}

\begin{IEEEbiography}[{\includegraphics
[width=1in,height=1.25in,keepaspectratio]{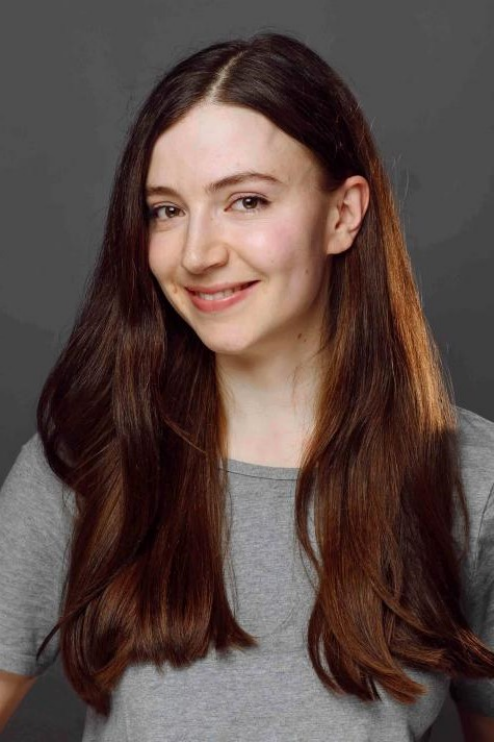}}]
{Ulla Scheler}
studied computer science at the Karlsruhe Institute of Technology (KIT) and the Technical University of Darmstadt. She received her Master's degree in Computer Science and joined MVTec Software GmbH as a research engineer in 2023.
\end{IEEEbiography}

\begin{IEEEbiography}[{\includegraphics
[width=1in,height=1.25in,clip,keepaspectratio]{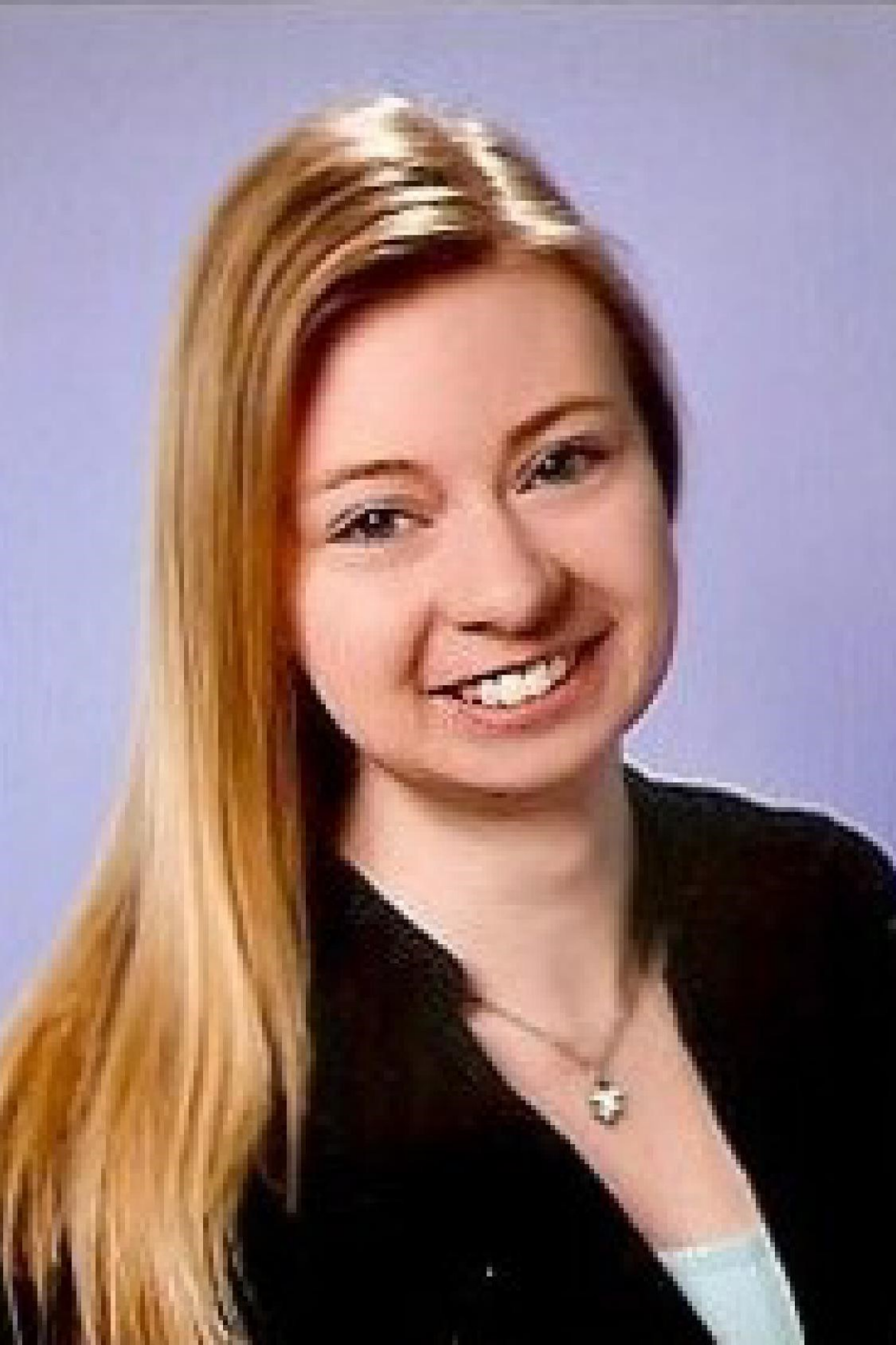}}]
{Rebecca König}
studied mathematics at the Technical University of Munich (TUM) and received her Master's degree in Mathematics in Science and Engineering from TUM in 2017. Since then, she has worked as a research engineer at MVTec Software GmbH. She has authored and coauthored multiple scientific publications in the field of deep learning in computer vision.
\end{IEEEbiography}

\begin{IEEEbiography}[{\includegraphics
[width=1in,height=1.25in,clip,
keepaspectratio]{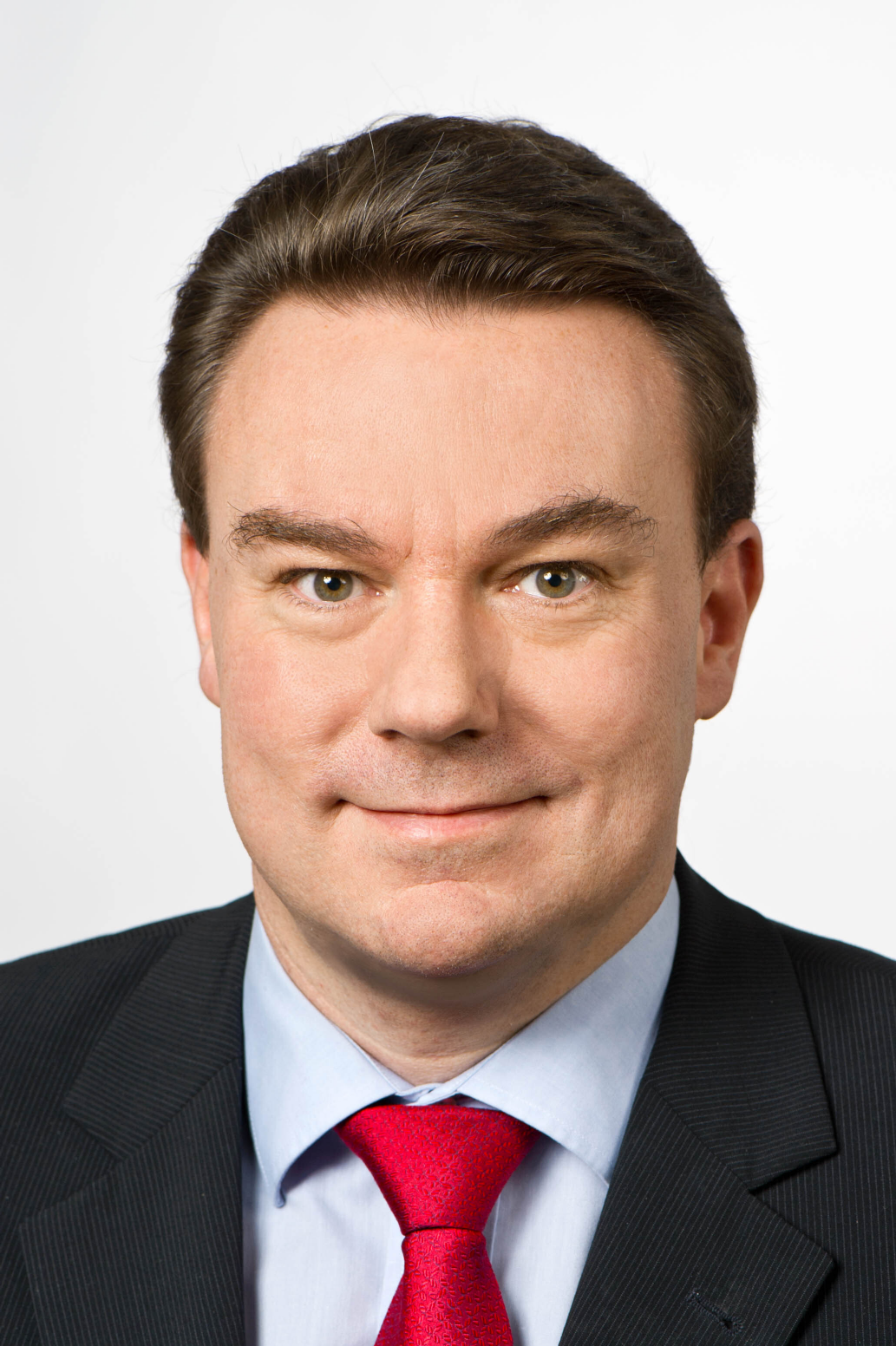}}]
{Carsten Steger}
studied computer science at the Technical University of Munich (TUM) and received the PhD degree from TUM in 1998. In 1996, he cofounded the company MVTec Software GmbH, where he heads the Research Department. He has authored and coauthored more than 100 scientific publications in the fields of computer and machine vision, including several textbooks on machine vision. In 2011, he was appointed a TUM honorary professor for the field of computer vision. He was a member of the Technical Committee of the German Association for Pattern Recognition (DAGM) from 2013 until 2021 and served as the spokesperson of the Technical Committee from 2018 until 2021.
\end{IEEEbiography}

\clearpage
\section*{Supplemental Material}
\renewcommand\appendixname{MVTec AD 2 - Supplemental Material}
\appendices


\section{Access to our dataset and evaluation server}
We provide all image data under \url{https://www.mvtec.com/company/research/datasets/mvtec-ad-2} together with ground truth annotations for \testpublic\ for local testing on a small scale.
Additionally, code snippets for measuring inference runtime and memory consumption can be downloaded along with general evaluation code and a PyTorch data loader for easy integration of \madtwo. 
The evaluation server is accessible via \url{https://benchmark.mvtec.com/}.

The dataset is published under the \href{https://creativecommons.org/licenses/by-nc-sa/4.0/}{CC BY-NC-SA 4.0 license} and hosted on \url{https://www.mvtec.com/company/research/datasets}. The linked server already hosts previous MVTec anomaly detection datasets, which are all still available in their entirety, and has proven itself as a stable solution for download availability and long-time preservation. In its structure, \madtwo\ follows the conventions and format set by previous anomaly detection datasets such as \mad\ \cite{bergmann2019_mvtec_ad_cvpr}. Because the dataset consists only of images, no special provisions must be made to read it.

\section{Additional Quantitative Results}
\subsection{Threshold-Independent Results}
As shown in the main paper, \aupro\ with a too high integration bound have little significance for datasets that contain small defects.
Nevertheless, to ensure comprehensive reporting, we include the \auprothirty\ results of the evaluated methods on \madtwo\ since this is currently the standard metric in the field of visual anomaly detection.
\Cref{tab:results_pro_03_256} presents the results for \testprivate\ and \testprivatemixed\ on an input image size of 256$\times$256.

In \Cref{fig:high_perf_at_higher_cost} of the main paper, we illustrated the trade-off between performance improvement and increased computational costs for the object \textit{Rice} when increasing the input image size of the anomaly detection models. 
\Cref{tab:results_pro_005_512} and \Cref{tab:results_pro_005_half_res} provide the \auprofive\ results for larger input dimensions on all objects in \madtwo. 
For example, PatchCore improves its performance from 28.8\%  on \testprivate\ (see \Cref{tab:results_aucpro_005} of the main paper) with input sizes of 256$\times$256 to 62.3\% when using half the original image width and height.
However, naively increasing the input image size to enhance performance can quickly exceed memory and runtime constraints of the anomaly detection system.
Furthermore, even with the highest tested image size, \madtwo\ leaves a significant room for improvement.

\begin{table*}[ht]
\centering
\caption{Anomaly segmentation \auprothirty\ performance (in \%) for input image size 256$\times$256 on \testprivate\ / \testprivatemixed.}
\label{tab:results_pro_03_256}
\begin{adjustbox}{width = .8\linewidth}
\begin{tabular}{lccccccc}
 \toprule
Object & PatchCore  & RD  & RD++ & EfficientAD & MSFlow & SimpleNet & DSR \\
\midrule
Can & 21.6 / 18.1 & 42.9 / 30.1 & 44.7 / 29.4 & 38.1 / 24.4 & 33.4 / 15.8 & 36.3 / 20.7 & 50.0 / 26.3 \\
Fabric & 34.6 / 35.3 & 22.3 / 25.6 & 26.2 / 29.5 & 46.8 / 41.3 & 38.2 / 38.4 & 25.7 / 26.0 & 23.7 / 25.6 \\
Fruit Jelly & 74.0 / 74.0 & 78.9 / 79.0 & 80.4 / 80.3 & 79.5 / 78.0 & 78.8 / 73.5 & 71.8 / 70.3 & 70.1 / 69.8 \\
Rice & 50.9 / 43.1 & 30.2 / 31.3 & 34.2 / 32.9 & 52.1 / 19.1 & 48.5 / 38.8 & 25.5 / 24.3 & 28.4 / 28.0 \\
Sheet Metal & 39.8 / 34.2 & 26.6 / 25.1 & 27.6 / 25.1 & 36.1 / 24.5 & 35.8 / 29.0 & 32.3 / 22.9 & 46.9 / 45.2 \\
Vial & 90.5 / 89.2 & 91.3 / 90.0 & 91.5 / 87.3 & 88.7 / 85.2 & 79.1 / 39.7 & 82.5 / 67.5 & 88.1 / 85.2 \\
Wall Plugs & 37.4 / 34.8 & 51.0 / 41.9 & 51.0 / 40.5 & 51.5 / 17.9 & 35.5 / 9.7 & 27.4 / 12.9 & 23.6 / 26.8 \\
Walnuts & 81.7 / 77.4 & 81.2 / 79.6 & 78.9 / 78.0 & 76.5 / 62.3 & 72.1 / 51.8 & 69.4 / 56.1 & 60.8 / 54.5 \\
\midrule
Mean & 53.8 / 50.8 & 53.0 / 50.3 & 54.3 / 50.4 & 58.7 / 44.1 & 52.7 / 37.1 & 46.4 / 37.6 & 49.0 / 45.2 \\

\bottomrule
\end{tabular}
\end{adjustbox}
\end{table*}

\begin{table*}[ht]
\centering
\caption{Anomaly segmentation \auprofive\ performance (in \%) for input image size 512$\times$512 on \testprivate\ / \testprivatemixed.} \label{tab:results_pro_005_512}
\begin{adjustbox}{width = .8\linewidth}
\begin{tabular}{lccccccc}
 \toprule
Object & PatchCore  & RD  & RD++ & EfficientAD & MSFlow & SimpleNet & DSR \\
\midrule
Can & 8.4 / 7.0 & 10.5 / 10.7 & 13.1 / 12.6 & 14.2 / 2.4 & 24.2 / 1.3 & 19.0 / 2.3 & 19.8 / 3.3 \\
Fabric & 23.7 / 23.5 & 36.8 / 39.6 & 41.2 / 43.0 & 44.5 / 28.3 & 31.7 / 28.2 & 22.5 / 23.6 & 30.3 / 28.7 \\
Fruit Jelly & 64.9 / 64.8 & 50.6 / 50.4 & 58.4 / 58.4 & 55.4 / 53.5 & 63.7 / 52.3 & 54.9 / 51.7 & 45.8 / 42.9 \\
Rice & 33.6 / 15.5 & 13.1 / 12.0 & 16.0 / 14.4 & 29.9 / 3.1 & 27.2 / 19.0 & 10.1 / 6.1 & 20.5 / 15.9 \\
Sheet Metal & 27.3 / 24.1 & 20.0 / 20.1 & 24.3 / 22.0 & 21.7 / 12.4 & 23.4 / 9.5 & 19.0 / 14.1 & 10.1 / 8.4 \\
Vial & 72.6 / 68.4 & 67.6 / 64.7 & 71.9 / 69.1 & 60.4 / 55.3 & 53.2 / 4.6 & 56.7 / 32.5 & 23.2 / 14.1 \\
Wall Plugs & 34.0 / 21.3 & 43.9 / 21.3 & 35.5 / 19.2 & 33.2 / 2.3 & 31.4 / 0.3 & 24.8 / 6.9 & 6.3 / 7.3 \\
Walnuts & 70.5 / 61.4 & 55.8 / 53.5 & 55.5 / 52.4 & 60.8 / 45.4 & 59.0 / 35.7 & 49.3 / 32.6 & 44.1 / 30.2 \\
\midrule
Mean & 41.9 / 35.8 & 37.3 / 34.0 & 39.5 / 36.4 & 40.0 / 25.3 & 39.2 / 18.9 & 32.0 / 21.2 & 25.0	/ 18.8 \\
\bottomrule
\end{tabular}
\end{adjustbox}
\end{table*}

\begin{table*}[ht]
\centering
\caption{Anomaly segmentation \auprofive\ performance (in \%) for half the original image width and height on \testprivate\ / \testprivatemixed. Due to the long training duration, DSR was only evaluated for object \textit{Rice} at this image size.} 
\label{tab:results_pro_005_half_res}
\begin{adjustbox}{width = .8\linewidth}
\begin{tabular}{lccccccc}
 \toprule
Object & PatchCore  & RD  & RD++ & EfficientAD & MSFlow & SimpleNet & DSR \\
\midrule
Can & 12.8 / 10.2 & 15.0 / 13.7 & 18.2 / 15.8 & 20.8 / 3.3 & 26.6 / 2.5 & 21.9 / 3.2 & - \\
Fabric & 69.0 / 59.1 & 81.1 / 77.9 & 82.1 / 77.8 & 72.5 / 45.4 & 82.9 / 60.6 & 66.0 / 55.6 & - \\
Fruit Jelly & 71.5 / 70.8 & 54.9 / 54.7 & 63.3 / 62.8 & 54.5 / 52.9 & 72.8 / 67.6 & 64.9 / 63.0 & - \\
Rice & 47.8 / 28.9 & 27.4 / 26.2 & 31.2 / 28.5 & 37.7 / 5.4 & 47.1 / 32.7 & 21.9 / 12.3 & 16.0 / 13.5 \\
Sheet Metal & 72.4 / 54.0 & 54.2 / 51.4 & 57.9 / 46.8 & 52.0 / 37.2 & 42.2 / 11.9 & 46.8 / 34.1 & - \\
Vial & 75.8 / 72.2 & 69.9 / 67.7 & 73.2 / 58.8 & 61.4 / 56.4 & 56.1 / 1.7 & 56.1 / 38.8 & - \\
Wall Plugs & 68.4 / 53.7 & 55.1 / 39.5 & 52.8 / 24.7 & 42.2 / 12.4 & 65.1 / 1.1 & 25.6 / 10.5 & - \\
Walnuts & 80.4 / 71.6 & 71.7 / 64.6 & 64.1 / 59.1 & 65.0 / 54.9 & 72.6 / 53.8 & 62.7 / 55.7 & - \\
\midrule
Mean & 62.3 / 52.6 & 53.7 / 49.5 & 55.3 / 46.8 & 50.8 / 33.5 & 58.2 / 29.0 & 45.7 / 34.1 & - \\
\bottomrule
\end{tabular}
\end{adjustbox}
\end{table*}

\subsection{Threshold-Dependent Results}
We base the image-level classification directly on the output of the anomaly detection methods, specifically the anomaly maps.
To determine whether an image is anomalous, we apply the segmentation threshold to these maps, computed as the mean plus three times the standard deviation of the anomaly-free validation images.
If at least one pixel exceeds this threshold and is thus classified as anomalous, the entire image is classified as \textit{reject}.
In \Cref{tab:results_class_f1}, we present the image-level classification $F_{1}$ score for \madtwo, evaluated on input image size of 256$\times$256. 
Interestingly, although PatchCore and MSFlow exhibit strong results for threshold-independent metrics, their classification performances lag behind those of the other investigated methods.
The poor performance of PatchCore can be attributed to its normalization procedure. 
By default, the anomaly maps are normalized across the entire input dataset to a range of $\left[0,1\right]$, which enables the combination of individual outputs from the ensemble models. 
This normalization approach works well when normal and anomalous images are processed together during inference. 
However, during the estimation of the segmentation threshold, only defect-free data is used as input. 
This causes the baseline threshold estimation method to derive an excessively high threshold, resulting in a significantly lower $F_{1}$ score compared to the other five best evaluated methods.

\begin{table*}[ht]
\centering
\caption{Classification $F_{1}$ score (in \%) on binarized images for \testprivate\ / \testprivatemixed\ set, respectively. Input image size is 256$\times$256. The segmentation threshold that decides whether a pixel is categorized as \textit{good} or \textit{bad} is determined on the validation set, i.e., using defect-free images only. The classification decision is based on whether at least one pixel is classified as anomalous.} \label{tab:results_class_f1}
\begin{adjustbox}{width = .8\linewidth}
\begin{tabular}{lccccccc}
 \toprule
Object & PatchCore  & RD  & RD++ & EfficientAD & MSFlow & SimpleNet & DSR \\
\midrule
Can & 41.4 / 57.8 & 50.9 / 66.5 & 51.1 / 65.9 & 57.5 / 67.1 & 15.0 / 64.4 & 48.2 / 67.0 & 70.8 / 70.8 \\
Fabric & 1.1 / 1.1 & 74.0 / 74.4 & 78.0 / 76.9 & 73.4 / 73.3 & 55.8 / 68.9 & 65.4 / 72.9 & 73.1 / 73.1 \\
Fruit Jelly & 9.3 / 8.3 & 84.1 / 84.1 & 85.8 / 85.2 & 84.2 / 84.2 & 64.1 / 78.6 & 84.6 / 84.3 & 83.8 / 83.8 \\
Rice & 2.2 / 2.2 & 78.8 / 78.8 & 78.0 / 78.5 & 79.2 / 79.2 & 53.5 / 77.5 & 69.8 / 76.3 & 79.0 / 79.0 \\
Sheet Metal & 1.9 / 1.9 & 43.4 / 44.4 & 35.8 / 43.2 & 86.4 / 85.1 & 65.1 / 74.9 & 82.0 / 83.4 & 85.5 / 85.5 \\
Vial & 4.0 / 4.0 & 83.3 / 83.5 & 83.7 / 83.5 & 83.5 / 83.5 & 75.6 / 81.8 & 84.8 / 83.7 & 83.5 / 83.5 \\
Wall Plugs & 1.4 / 0.0 & 73.9 / 73.9 & 72.2 / 73.4 & 74.0 / 73.6 & 4.2 / 53.4 & 69.1 / 73.4 & 74.1 / 73.9 \\
Walnuts & 2.9 / 4.3 & 78.2 / 77.3 & 79.5 / 77.2 & 76.1 / 76.1 & 75.6 / 76.6 & 82.6 / 76.3 & 74.4 / 74.4 \\
\midrule
Mean & 8.0 / 9.9 & 70.8 / 72.9 & 70.5 / 73.0 & 76.8 / 77.8 & 51.1 / 72.0 & 73.3 / 77.1 & 78.0 / 78.0 \\
\bottomrule
\end{tabular}
\end{adjustbox}
\end{table*}

\section{Example Defects}
\begin{figure*}[ht]
\centering
\begin{adjustbox}{max width = .98\linewidth}
  \includegraphics{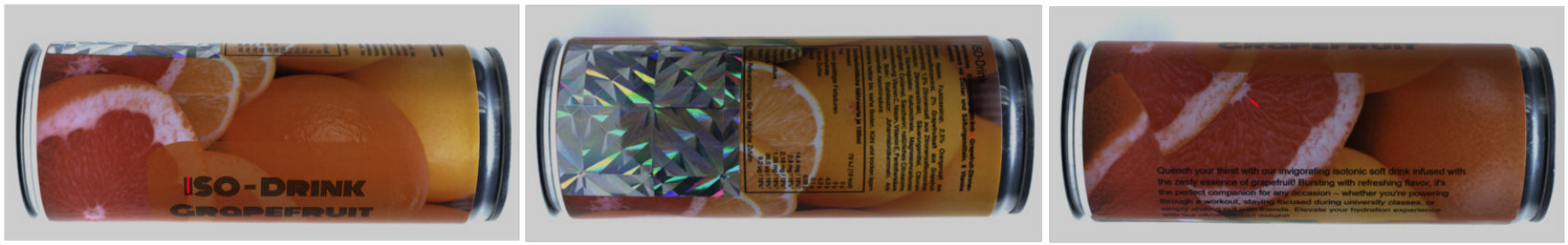}
\end{adjustbox}
\caption{Example images from the \testpublic\ set for \textit{Can}. The can is randomly rotated and contains a glitter foil to cover barcode information, which adds additional allowable variation. The defects depicted here are print errors, changes in color, and scratches. Other defects include missing can labels.}
\label{fig:example_defects_can}
\end{figure*}

\begin{figure*}[ht]
\centering
\begin{adjustbox}{max width = .98\linewidth}
  \includegraphics{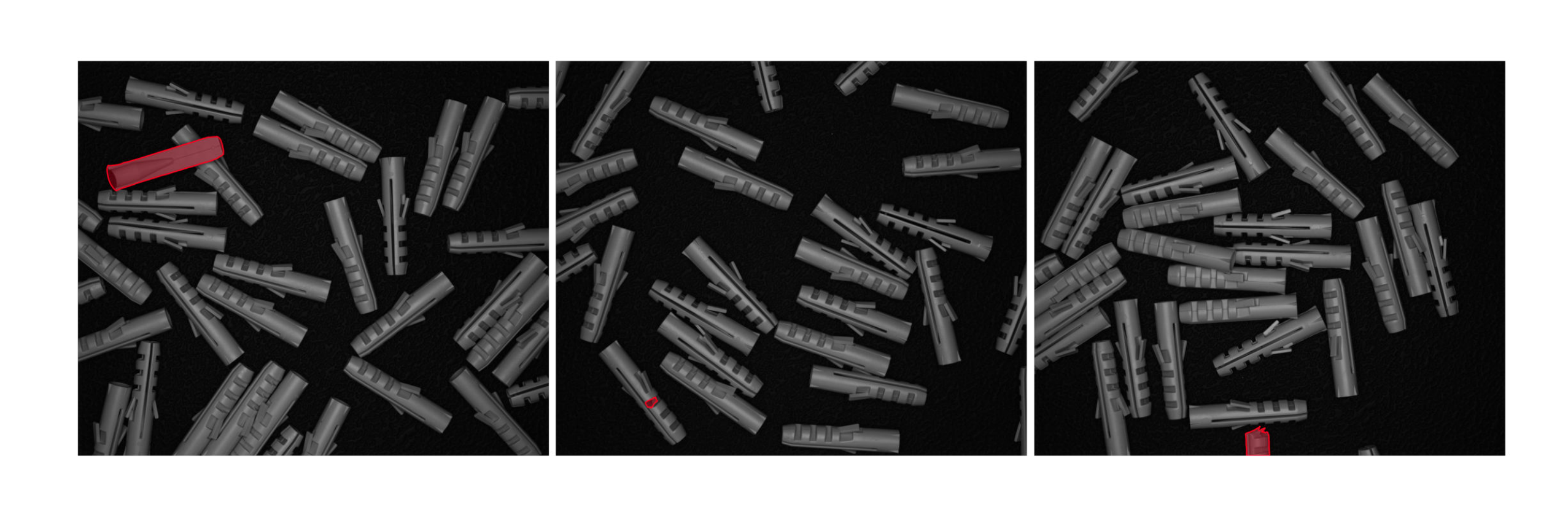}
\end{adjustbox}
\caption{Example images from the \testpublic\ set for \textit{Wall Plugs}. The defects depicted here are cuts, scratches, contaminations, and broken object instances. Other defects include bent and missing parts as well as wall plugs of the wrong size. Wall plugs can vary in their quantity and positions and may overlap, leading to occlusions. Defects occur all over the image, including close to the image borders.}
\label{fig:example_defects_wall_plugs}
\end{figure*}

\begin{figure*}[ht]
\centering
\begin{adjustbox}{max width = .98\linewidth}
  \includegraphics{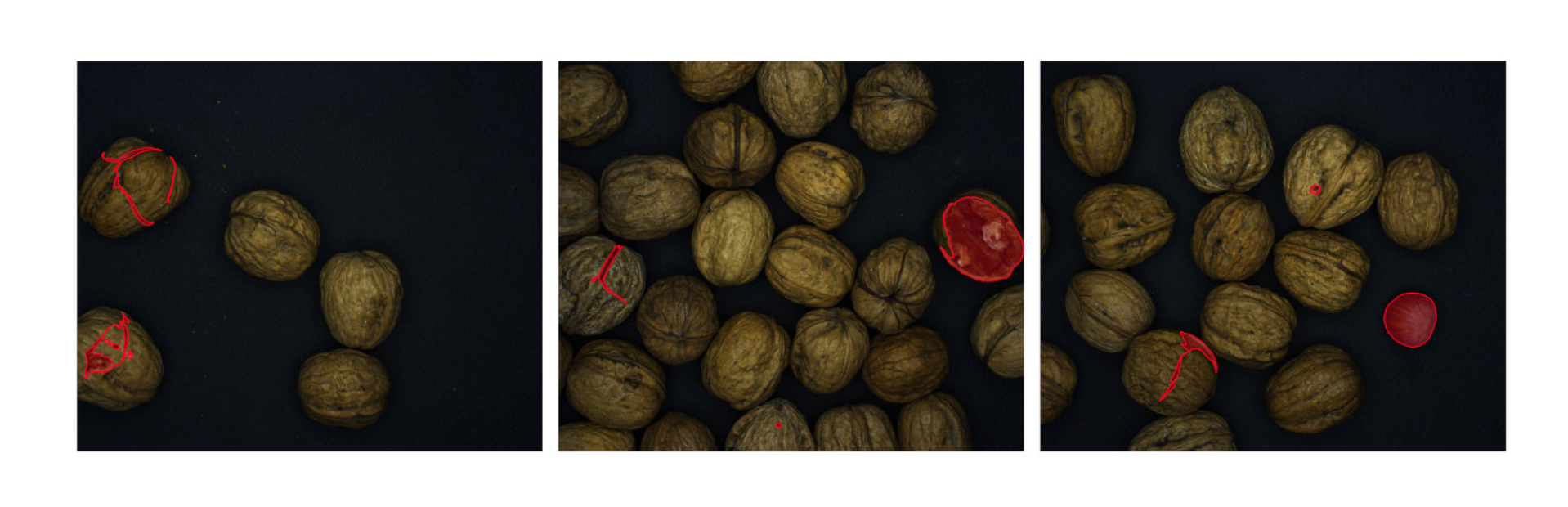}
\end{adjustbox}
\caption{Example images from the \testpublic\ set for \textit{Walnuts}. The defects depicted here are cracks, holes, and contamination by foreign bodies, in this case a hazelnut. Similar to the \textit{Wall Plugs} scenario, walnuts appear in random positions and quantity. Additionally, walnuts exhibit large natural variations in their color, shape, and size. Defects occur all over the image, including close to the image borders.}
\label{fig:example_defects_walnuts}
\end{figure*}

\begin{figure*}[ht]
\centering
\begin{adjustbox}{max width = .98\linewidth}
  \includegraphics{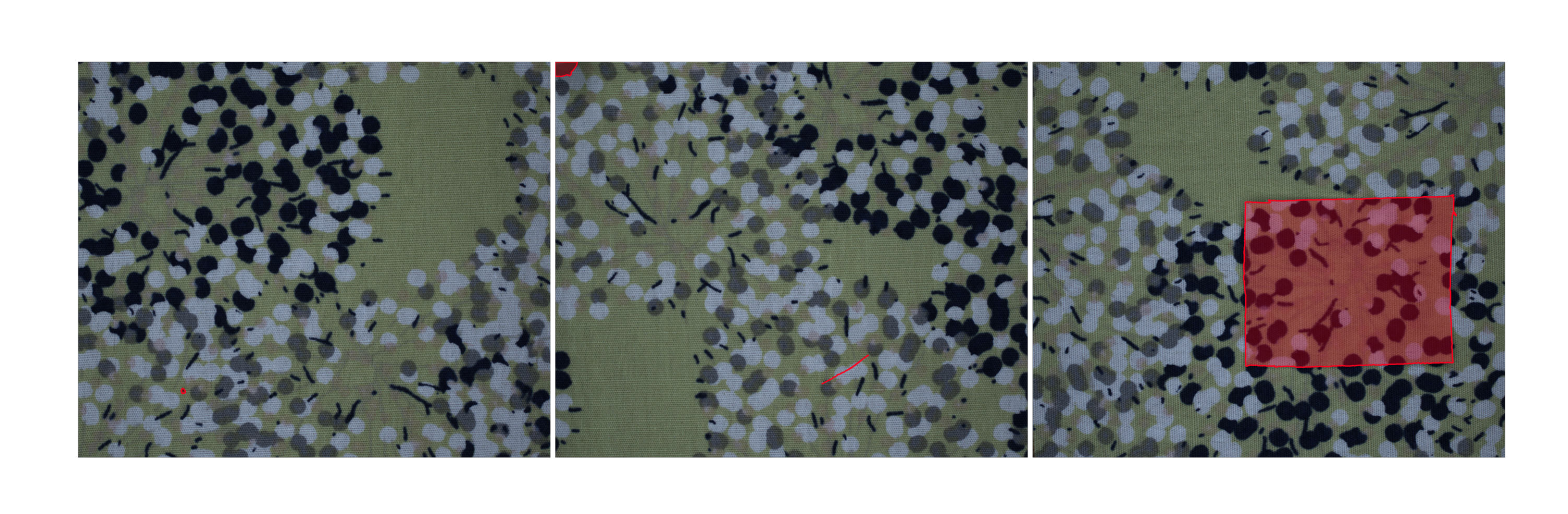}
\end{adjustbox}
\caption{Example images from the \testpublic\ set for \textit{Fabric}. The pattern is variable and varies in orientation. The defects depicted here are cuts, holes, and extraneous pieces of fabric. Other defects include discoloration and additional yarn. Defects occur all over the image, including close to the image borders.}
\label{fig:example_defects_fabric}
\end{figure*}

\begin{figure*}[p]
\centering
\begin{adjustbox}{max width = .98\linewidth}
  \includegraphics{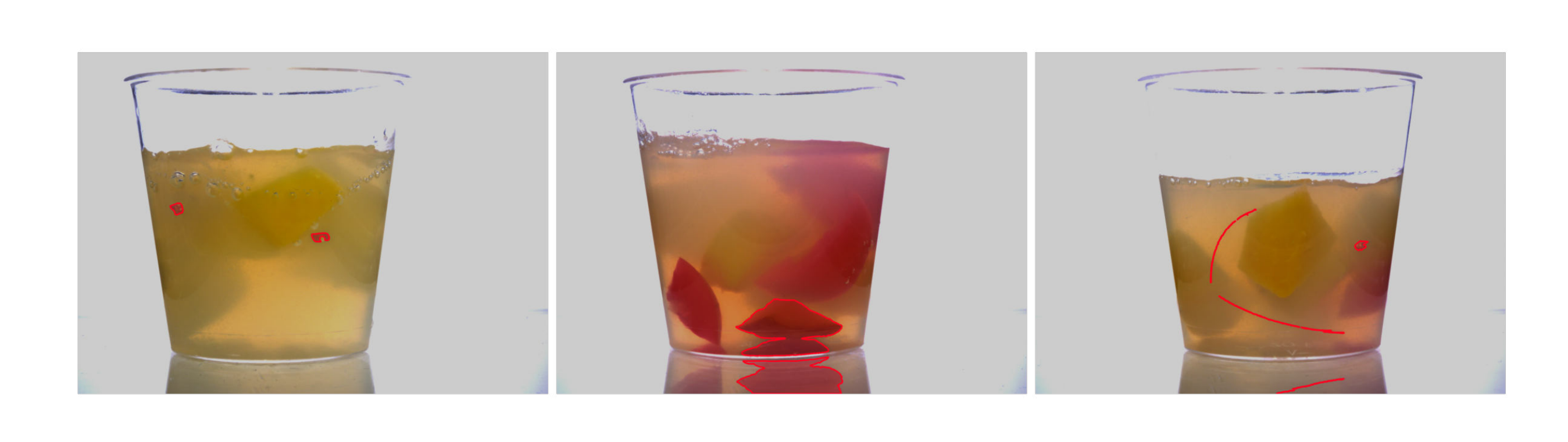}
\end{adjustbox}
\caption{Example images from the \testpublic\ set for \textit{Fruit Jelly}. The defects depicted here are alien objects like metal flakes and vegetable pieces as well as hairs. Other defects include further types of contaminations or scratches in the glass. For completeness sake, the reflection of the defects on the bottom of the setup were labeled as well. Differently colored fruits, the refractive properties of the jelly, and the random combination of ingredients lead to a large variety in the non-anomalous data.}
\label{fig:example_defects_fruit_jelly}
\end{figure*}

\begin{figure*}[ht]
\centering
\begin{adjustbox}{max width = .98\linewidth}
  \includegraphics{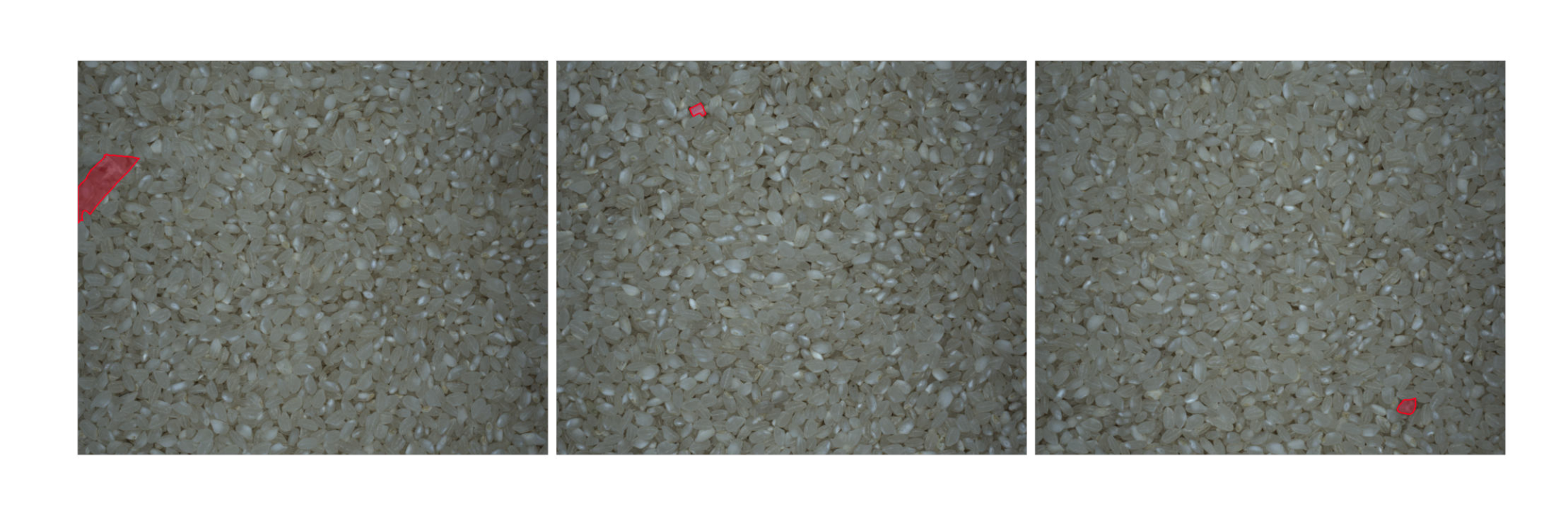}
\end{adjustbox}
\caption{Example images from the \testpublic\ set for \textit{Rice}. The defects depicted here are plastic foil and foreign bodies that were chosen to be visually similar to rice in shape, size, and color. Other defects include hairs. Defects occur all over the image, including close to the image borders.}
\label{fig:example_defects_rice}
\end{figure*}

\begin{figure*}[ht]
\centering
\begin{adjustbox}{max width = .98\linewidth}
  \includegraphics{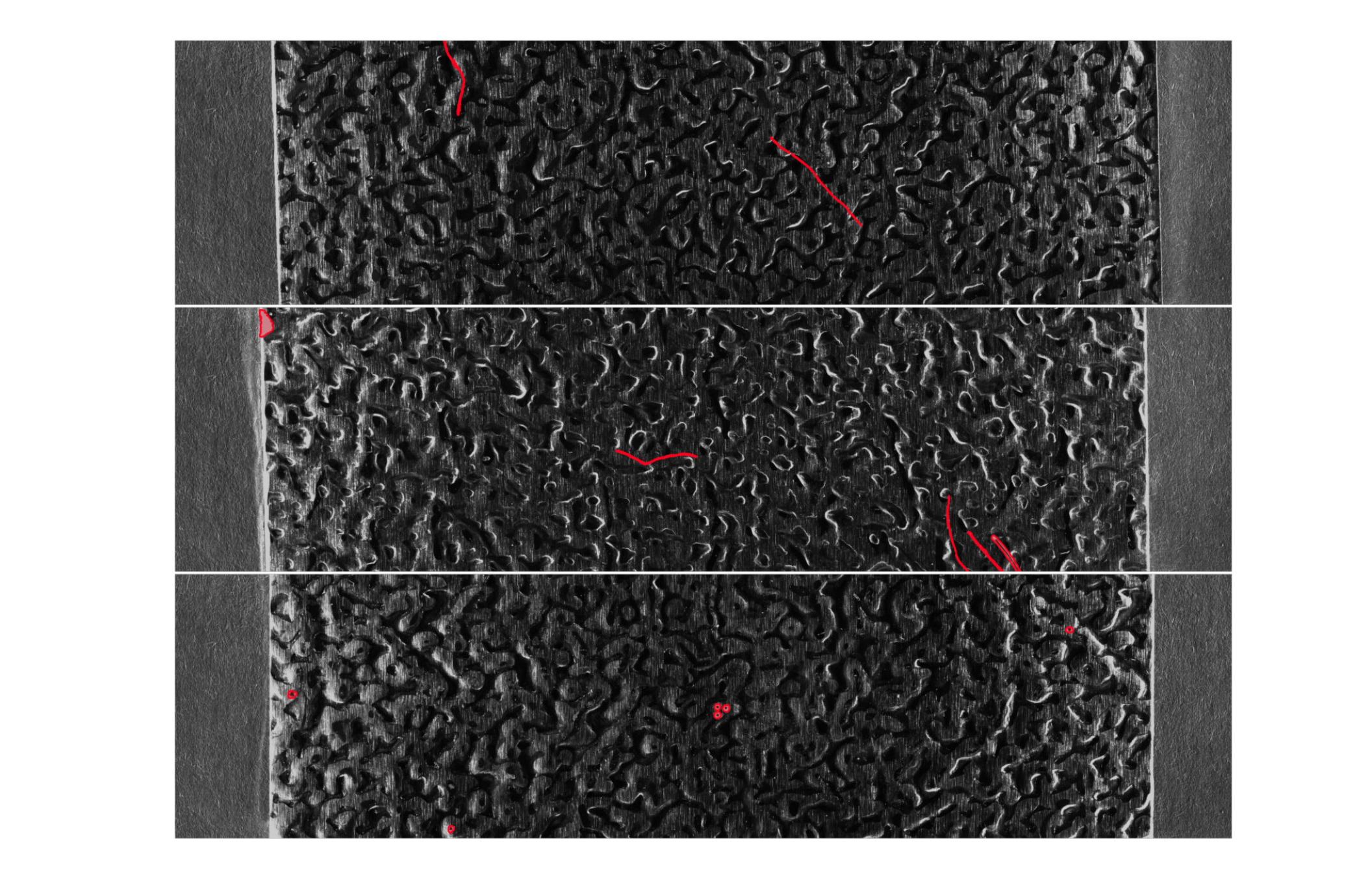}
\end{adjustbox}
\caption{Example images from the \testpublic\ set for \textit{Sheet Metal}. The defects depicted here are scratches, cuts, and holes. Other defects include foreign bodies covering the material. Sheet metal exhibits a large variation in the non-anomalous data, intensified by specular highlights and shadows. Defects occur all over the image, including close to the image borders.}
\label{fig:example_defects_sheet_metal}
\end{figure*}

\begin{figure*}[ht]
\centering
\begin{adjustbox}{max width = .98\linewidth}
  \includegraphics{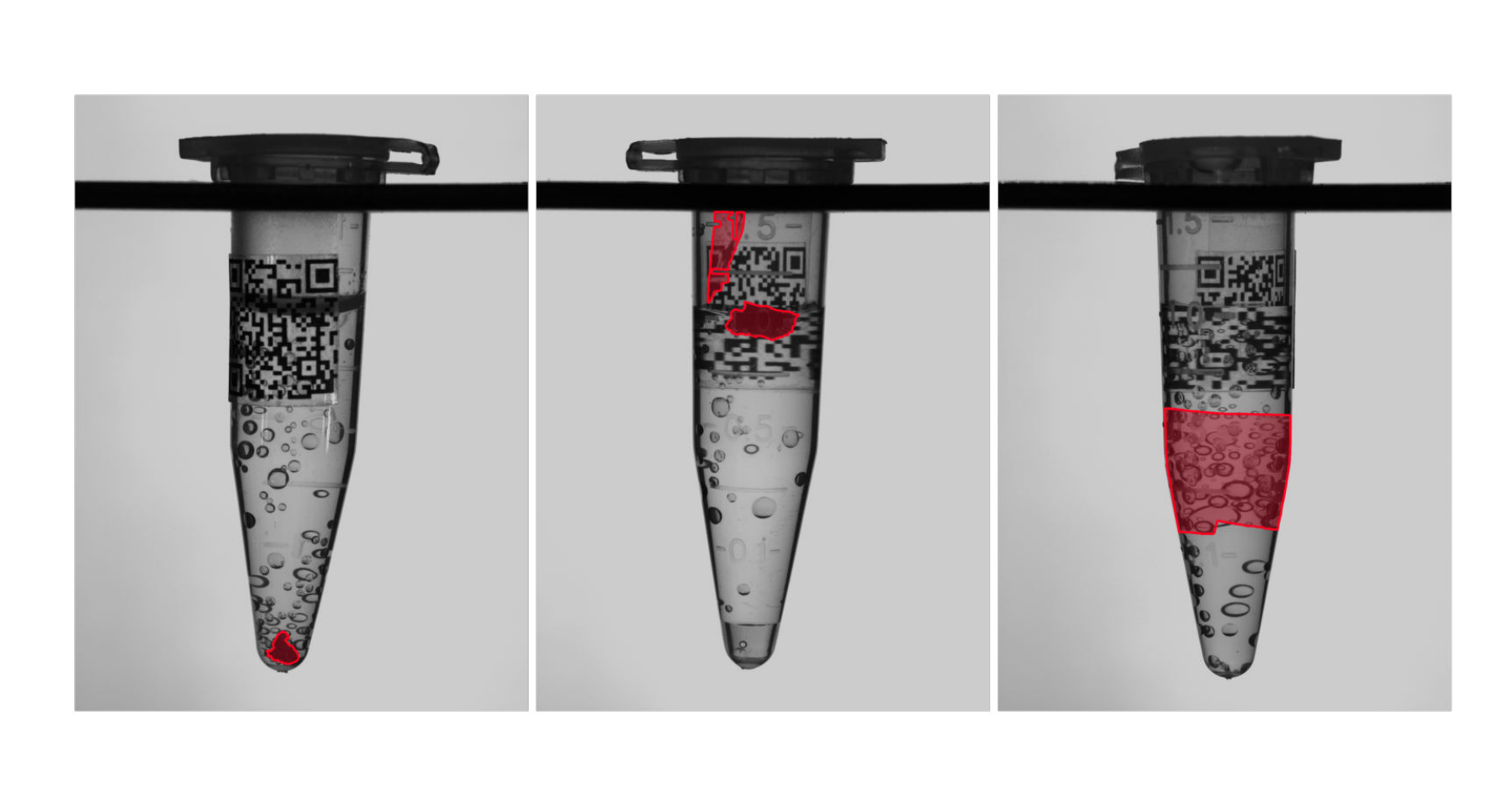}
\end{adjustbox}
\caption{Example images from the \testpublic\ set for \textit{Vial}. The defects depicted here are foreign bodies and plastic foil. Other defects include additional or missing QR codes, too much or too little liquid, hairs, and open or missing lids. QR codes, air bubbles, rotations of the vial, and distortions caused by the liquid lead to a large variations in the non-anomalous data.}
\label{fig:example_defects_vial}
\end{figure*}

\Crefrange{fig:example_defects_can}{fig:example_defects_vial} provide images with examples of defects for each object category from \madtwo.

\section{Qualitative Results}
\begin{figure*}[ht]
\centering
\begin{adjustbox}{max width = .98\linewidth}
  \includegraphics[]{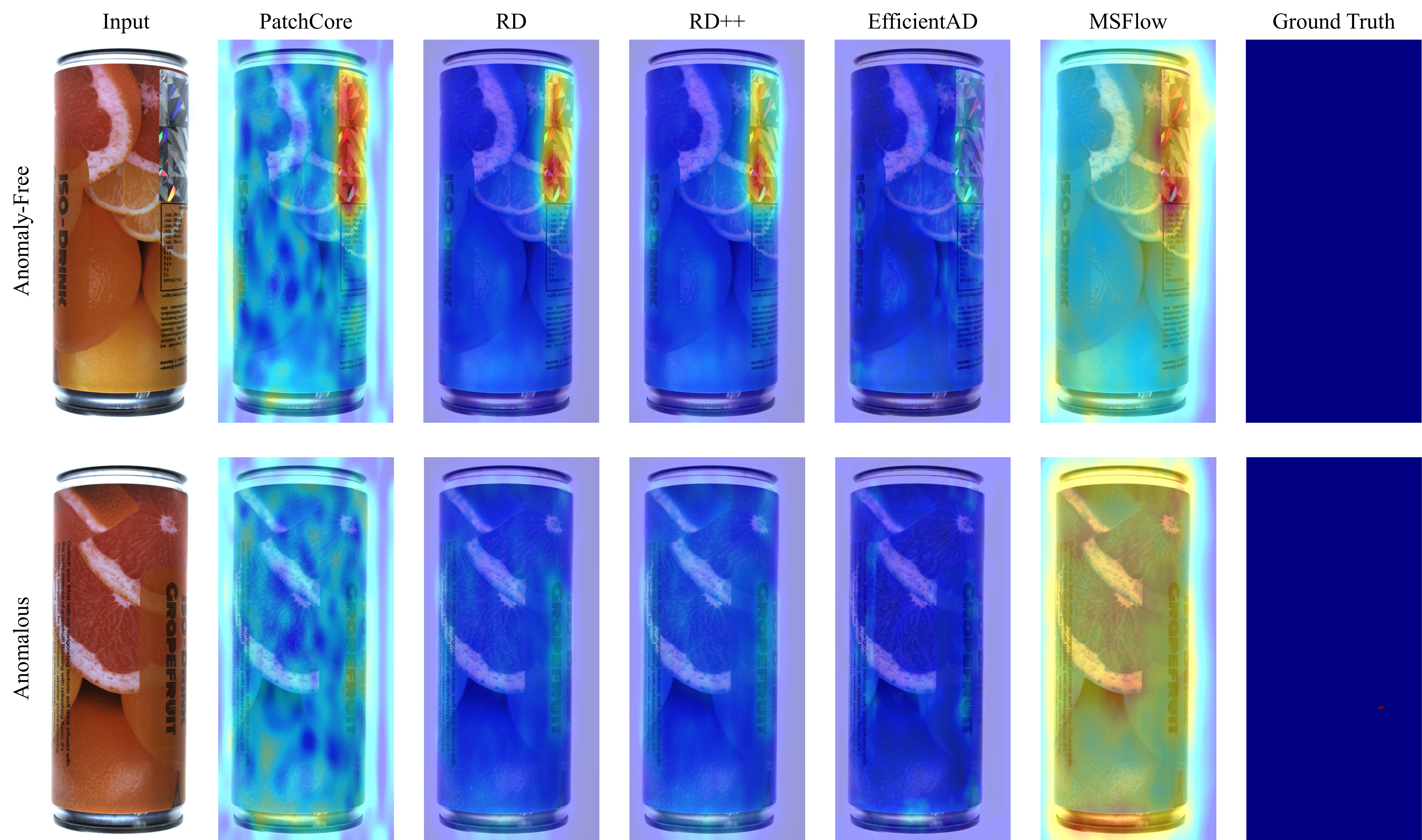}
\end{adjustbox}
\caption{Anomaly-free and anomalous examples of the \madtwo\ object \textit{Can} and the respective output maps of the five best evaluated methods. Ground truth of anomalous image best viewed with zoom.}
\label{fig:ad_maps_can}
\end{figure*}

\begin{figure*}[ht]
\centering
\begin{adjustbox}{max width = .98\linewidth}
  \includegraphics[]{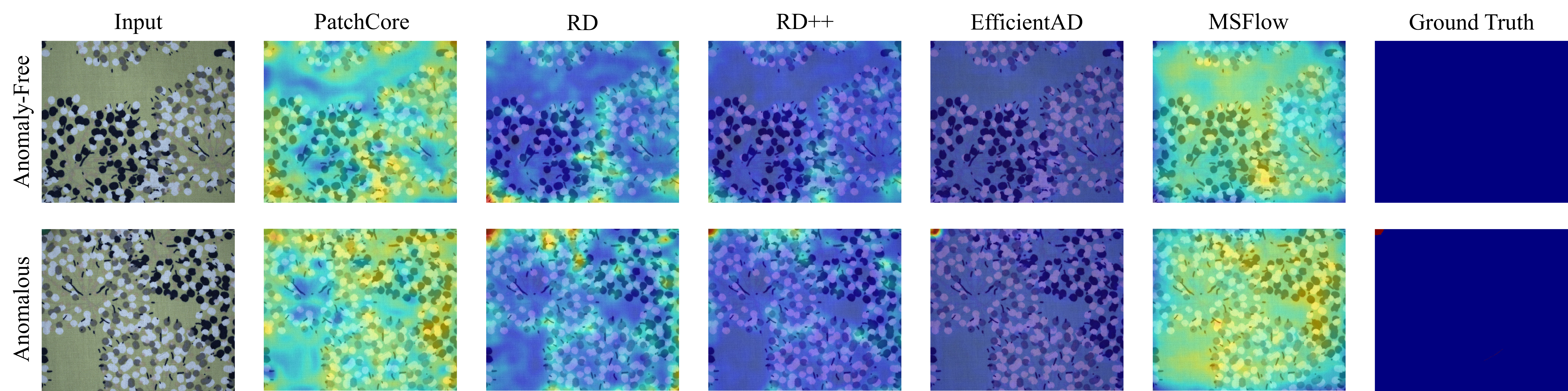}
\end{adjustbox}
\caption{Anomaly-free and anomalous examples of the \madtwo\ object \textit{Fabric} and the respective output maps of the five best evaluated methods. Ground truth of anomalous image best viewed with zoom.}
\label{fig:ad_maps_fabric}
\end{figure*}

\begin{figure*}[ht]
\centering
\begin{adjustbox}{max width = .98\linewidth}
  \includegraphics[]{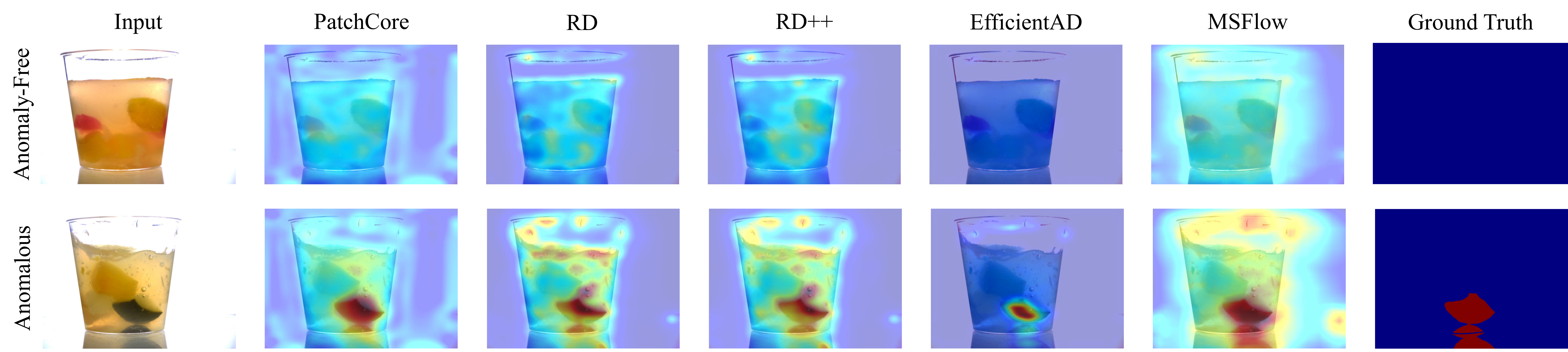}
\end{adjustbox}
\caption{Anomaly-free and anomalous examples of the \madtwo\ object \textit{Fruit Jelly} and the respective output maps of the five best five best evaluated methods.}
\label{fig:ad_maps_fruit_jelly}
\end{figure*}

\begin{figure*}[ht]
\centering
\begin{adjustbox}{max width = .98\linewidth}
  \includegraphics[]{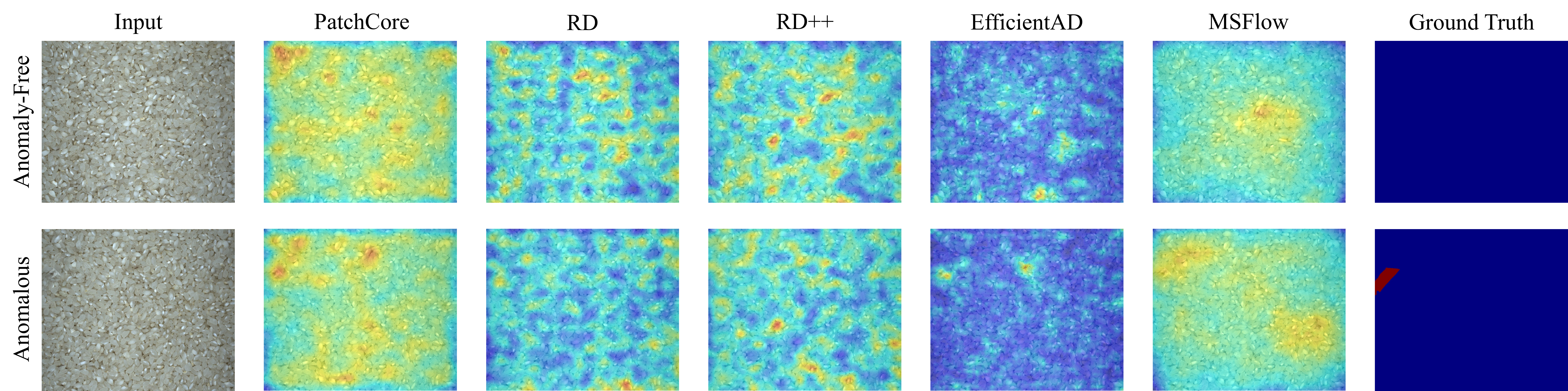}
\end{adjustbox}
\caption{Anomaly-free and anomalous examples of the \madtwo\ object \textit{Rice} and the respective output maps of the five best evaluated methods.}
\label{fig:ad_maps_rice}
\end{figure*}

\begin{figure*}[ht]
\centering
\begin{adjustbox}{max width = .98\linewidth}
  \includegraphics[]{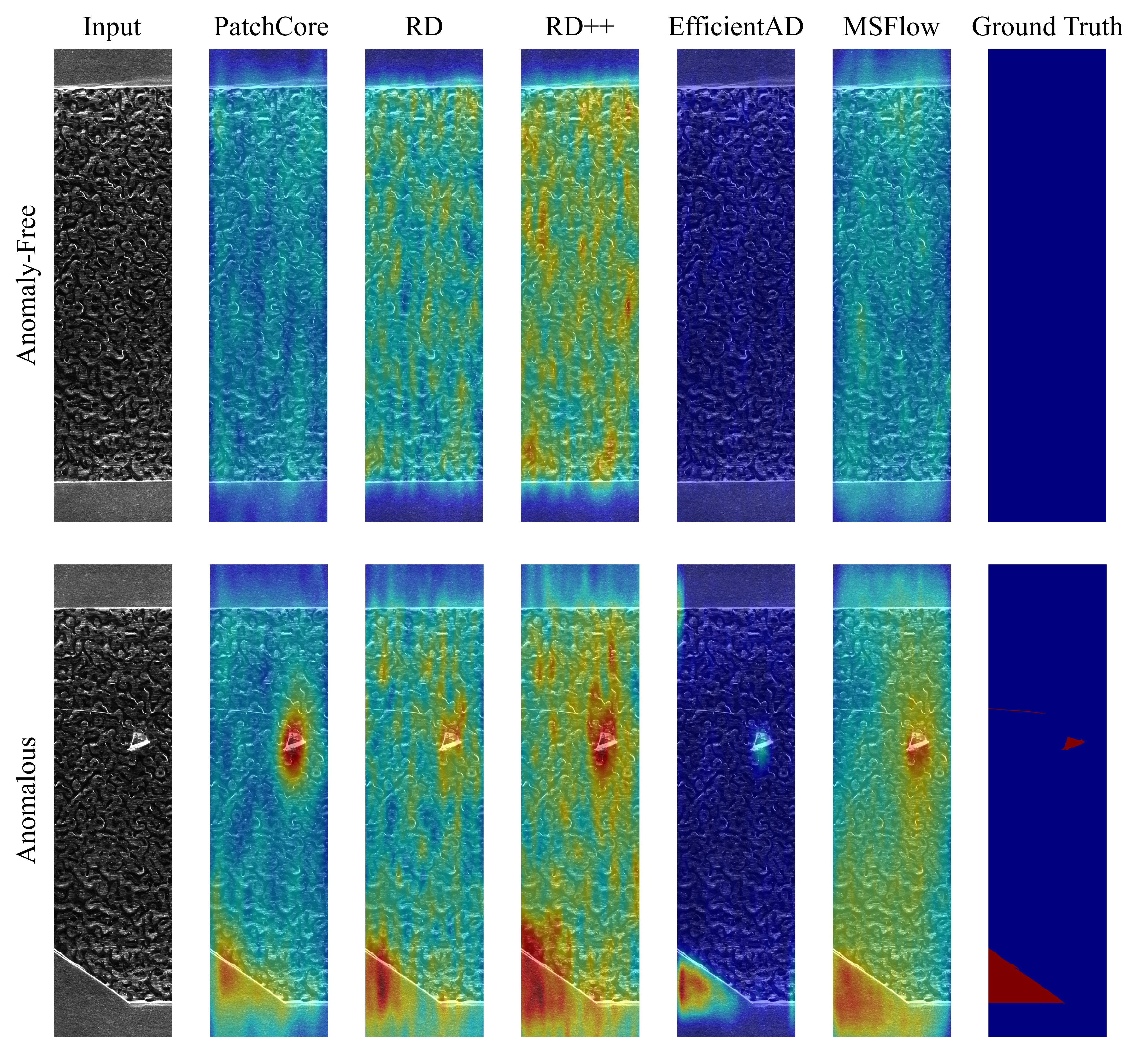}
\end{adjustbox}
\caption{Anomaly-free and anomalous examples of the \madtwo\ object \textit{Sheet Metal} and the respective output maps of the five best evaluated methods.}
\label{fig:ad_maps_sheet_metal}
\end{figure*}

\begin{figure*}[ht]
\centering
\begin{adjustbox}{max width = .98\linewidth}
  \includegraphics[]{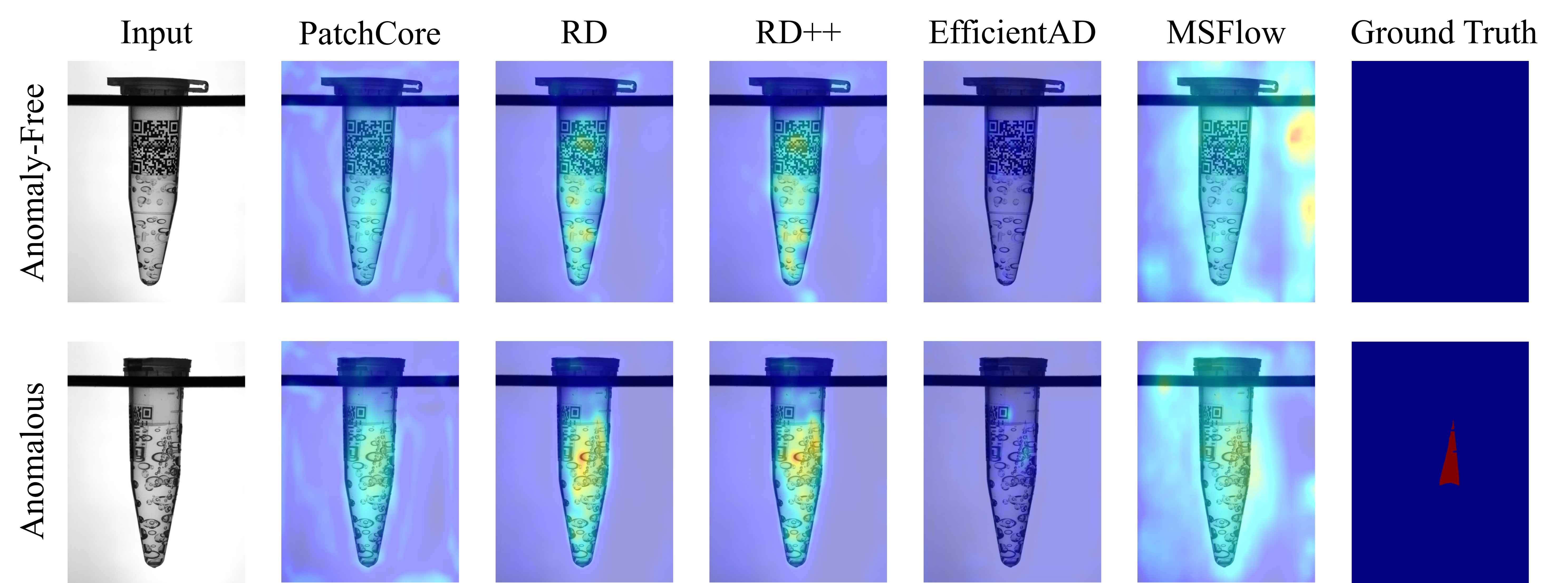}
\end{adjustbox}
\caption{Anomaly-free and anomalous examples of the \madtwo\ object \textit{Vial} and the respective output maps of the five best evaluated methods.}
\label{fig:ad_maps_vial}
\end{figure*}

\begin{figure*}[ht]
\centering
\begin{adjustbox}{max width = .98\linewidth}
  \includegraphics[]{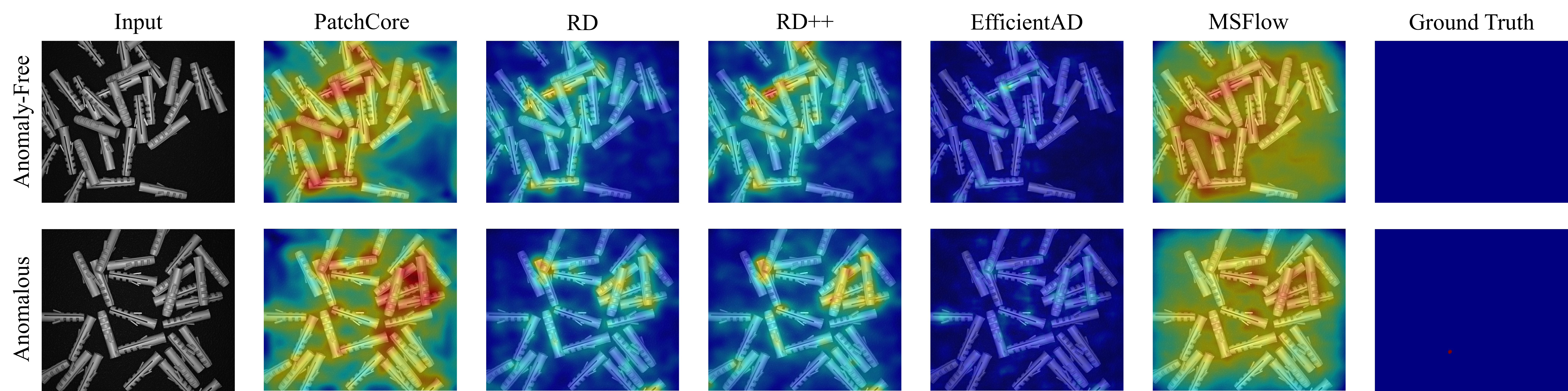}
\end{adjustbox}
\caption{Anomaly-free and anomalous examples of the \madtwo\ object \textit{Wall Plugs} and the respective output maps of the five best evaluated methods.}
\label{fig:ad_maps_wallplugs}
\end{figure*}

\begin{figure*}[ht]
\centering
\begin{adjustbox}{max width = .98\linewidth}
  \includegraphics[]{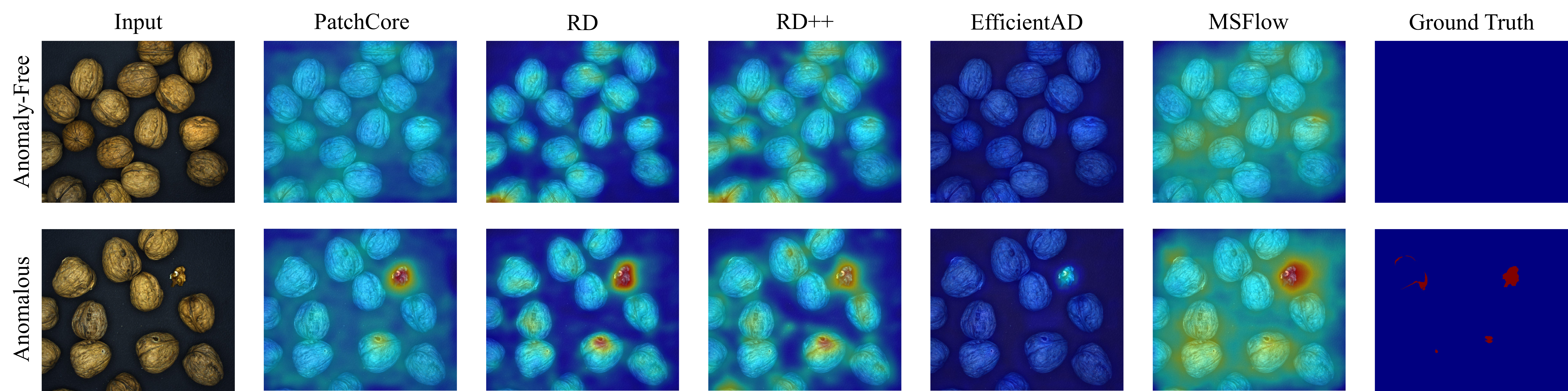}
\end{adjustbox}
\caption{Anomaly-free and anomalous examples of the \madtwo\ object \textit{Walnuts} and the respective output maps of the five best evaluated methods.}
\label{fig:ad_maps_walnuts}
\end{figure*}

In \Crefrange{fig:ad_maps_can}{fig:ad_maps_walnuts}, we present both anomaly-free and anomalous example images of the \madtwo\ objects.
Additionally, we provide anomaly output maps for the five best evaluated methods with an input size of 256$\times$256.
For better visualization, the colormap limits are fixed to the minimum and maximum pixel anomaly scores of the methods' outputs across \testpublic.

\end{document}